\documentclass[preprint,times,authoryear]{elsarticle}

\usepackage[T1]{fontenc}

\usepackage{amssymb}
\usepackage{amsmath}

\usepackage{booktabs}           %
\usepackage{multirow}           %
\usepackage{array}              %

\usepackage{tikz}               %
\usetikzlibrary{positioning, arrows.meta, calc}
\usepackage{subcaption}         %

\usepackage{placeins}           %
\usepackage{float}              %
\usepackage{setspace}           %

\AtBeginEnvironment{table}{\singlespacing}
\AtBeginEnvironment{tabular}{\singlespacing}
\AtBeginEnvironment{figure}{\singlespacing}

\usepackage{xspace}             %
\usepackage{xcolor}             %
\usepackage{colortbl}           %

\usepackage[breaklinks,colorlinks,allcolors=blue]{hyperref}

\journal{Expert Systems with Applications}

\begin{document}

\begin{frontmatter}

\title{Robust Logical Anomaly Detection under Capture Shifts via Synthetic Anomalies}
\author[metu]{Ufuk Efe\corref{cor1}}
\ead{ufuk.efe@metu.edu.tr}

\author[metu]{Alptekin Temizel}
\ead{atemizel@metu.edu.tr}

\cortext[cor1]{Corresponding author.}

\affiliation[metu]{organization={Graduate School of Informatics, Middle East Technical University},
            addressline={\"{U}niversiteler Mahallesi, Dumlup\i{}nar Bulvar\i{} No:1},
            city={Ankara},
            postcode={06800},
            country={T\"{u}rkiye}}

\begin{abstract}

Detectors of logical anomalies, such as missing, extra or wrongly combined parts, usually score a test image by its distance from normal (good) images. They achieve near-perfect image-level AUROC on standard industrial anomaly detection benchmarks such as MVTec LOCO, but their robustness to changes in capture conditions remains largely unexamined. Under a heterogeneous perturbation protocol in which each test image is randomly kept unaltered or subjected to a brightness reduction or a zoom-out transformation, the normal-reference methods we tested lose 9.7 to 34.1 AUROC points, and two of the strongest, CSAD and SALAD, fall from 97.7\% and 96.4\% to 68.1\% and 62.3\%. We propose a framework in which a classifier, trained on normal images and on synthetic anomalies, scores a test image without comparing it with normal data. From a short description of each product, a language model derives anomaly types and editing instructions, an image editor applies them to normal images, and the classifier learns to separate the edited images from automatically discovered normal sub-modes. With its full training recipe, the framework reaches 94.9\% AUROC on clean images, 1.5 to 2.8 points below SALAD and CSAD, and 94.1\% under the perturbation protocol. Trained without brightness or zoom-out augmentation, so that neither test perturbation occurs in training, it still reaches 92.5\% (93.2\% clean). These results suggest that clean benchmark scores can hide a strong sensitivity to capture conditions, and that learning the decision from both normal images and rule violations may reduce it.
\end{abstract}


\end{frontmatter}

\section{Introduction}
\label{sec:intro}

\begin{figure}[H]
    \centering
    \includegraphics[width=0.8\linewidth]{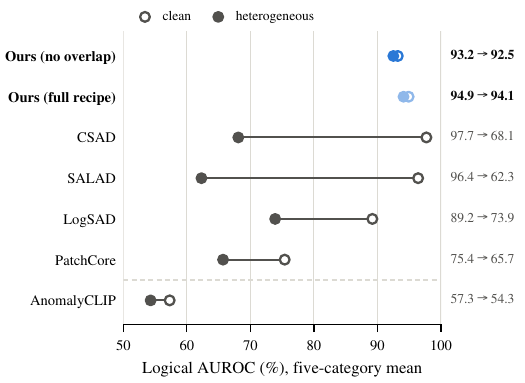}
    \caption{\textbf{Clean and heterogeneous AUROC.} Five-category mean image-level logical AUROC on MVTec LOCO for clean test images and under the heterogeneous perturbation protocol, in which each test image is randomly kept unaltered or subjected to a brightness reduction or a zoom-out transformation. Our two variants are shown in blue: Ours (full recipe) and Ours (no overlap), which is trained without brightness or zoom-out augmentation and is therefore used to assess robustness. AnomalyCLIP (below the dashed line), a zero-shot method that uses no normal reference, is near chance level even on clean images.}
    \label{fig:teaser}
\end{figure}

Visual anomaly detection is essential for automated quality control in manufacturing~\citep{lin2025survey}. Because defective products are rare and diverse, most detectors are trained exclusively on normal images, without access to real anomalies~\citep{lin2025survey,bergmann2022beyond}. We study \emph{logical} anomalies, such as missing, extra or wrongly combined parts, rather than structural defects such as scratches. On MVTec LOCO~\citep{bergmann2022beyond}, the standard benchmark for this task, the best methods, such as CSAD~\citep{hsieh2024csad} and SALAD~\citep{fucka2025salad}, reach near-perfect image-level area under the ROC curve (AUROC). Although both also train on synthetic anomalies, their score still depends on how far a test image lies from normal data; we call such methods \emph{normal-reference} methods. These benchmark results, however, come from images taken under controlled capture conditions.

On a production line, illumination and camera position do not always remain constant~\citep{baitieva2025beyond}. Robustness to such capture shifts has been benchmarked mainly for structural defects~\citep{cao2023anomaly,pemula2025robustad,heckler2025mvtecad2}. We evaluate logical anomaly detectors on MVTec LOCO under a heterogeneous perturbation protocol (Section~\ref{sec:evaluation}). Each test image is assigned at random to one of seven conditions, unperturbed or with a brightness reduction or a zoom-out at one of three strengths, and a single AUROC is computed over the mixed test set. Under this protocol, the normal-reference methods we tested lose 9.7 to 34.1 AUROC points, and CSAD and SALAD fall to 68.1\% and 62.3\% (Fig.~\ref{fig:teaser}). Since neither perturbation alters the logical content of an image, this drop originates in how the detectors score the perturbed images.

A likely reason for this drop is that normal-reference detectors compare a test image with normal data taken under one lighting and one camera position. A darker image or a smaller object may then be scored as anomalous even when every part is correct; indeed, under both perturbations, many normal test images score above the median clean anomaly (Fig.~\ref{fig:score_shift}). This reading suggests learning the decision instead as a contrast between normal images and rule violations seen under the same capture conditions, so that a capture change moves both sides alike. Our framework differs from CSAD and SALAD not in using synthetic anomalies but in this test-time decision, which is \emph{fully discriminative}: the classifier scores a test image without comparing it with normal data. Learning this contrast, however, needs rule violations, and real ones are rarely available.

Our framework therefore creates these rule violations as synthetic anomalies from a description of the product, so that, like the normal-reference methods, it needs no real anomaly during training. First, a person writes a short description of each category. From it, a language model proposes anomaly types, rule violations that could occur in the category, and writes an editing prompt for each; an image editor then applies one prompt to each normal training image. The edited images are then manually reviewed, assisted by a vision-language model, and only valid anomalies are retained. Normal sub-mode discovery, a simple automatic step, assigns each distinct group of normal images its own class, and the classifier is trained to distinguish these normal sub-modes from the anomaly types. Similar generate-then-train pipelines exist for structural defects~\citep{sun2025anomalyanything,fucka2026anomalyvfm,hu2026mirage}; for logical anomalies, ComGEN~\citep{tong2025comgen} generates anomalies with a diffusion model but keeps a normal-reference decision. In summary, our contributions are threefold:
\begin{enumerate}
    \item We show that brightness reduction and zoom-out, applied at random to the test images of MVTec LOCO, reduce the AUROC of the normal-reference methods we tested by 9.7 to 34.1 points, a loss of 38.2\% to 73.5\% of their performance above chance (Section~\ref{sec:evaluation}).
    \item We propose a framework in which a classifier, trained on normal images and on synthetic anomalies derived from a short description of each category, scores a test image without comparing it with normal data.
    \item With its full training recipe, the framework reaches 94.9\% AUROC on clean images, 1.5 to 2.8 points below SALAD and CSAD, and 94.1\% under the same perturbations; trained without brightness or zoom-out augmentation, it still reaches 92.5\% (93.2\% clean).
\end{enumerate}
\section{Related Work}
\label{sec:related}

\subsection{Logical Anomaly Detection}
\label{sec:related_logical}

MVTec LOCO~\citep{bergmann2022beyond} is the standard benchmark for logical anomalies, which break rules on the presence, number, position or combination of parts, unlike the local structural defects of MVTec AD~\citep{bergmann2019mvtec}. Because such a rule spans the whole image, the strongest methods model the image's composition. CSAD~\citep{hsieh2024csad} segments components without manual labels and compares their patch histograms with those of normal images, and SALAD~\citep{fucka2025salad} adds a composition branch, trained on synthetic anomalies, to its appearance branches. PSAD~\citep{kim2024psad} learns part segmentation from a few labeled images, and SAM-LAD~\citep{peng2025samlad} matches segmented objects with those of retrieved normal images. LogSAD~\citep{zhang2025logsad} needs no training: it compares a test image with a memory of normal features and checks matching rules that its authors built from GPT-4V proposals, prompted with a few normal images and the benchmark's constraint descriptions. By contrast, general-purpose detectors such as PatchCore~\citep{roth2022patchcore} model local normality and struggle with logical anomalies~\citep{kim2024psad}, and zero-shot detectors such as AnomalyCLIP~\citep{zhou2024anomalyclip} use no image of the target category. Apart from AnomalyCLIP, all methods above are normal-reference methods: they score a test image, at least in part, by its distance from normal training data. We examine this normal-reference design under capture shifts (Section~\ref{sec:experiments}).

Another line of work employs language models for detection. LogicAD~\citep{jin2025logicad} describes images with text features from a vision-language model (VLM), and LogiCode~\citep{zhang2025logicode} uses a large language model (LLM) to generate code that checks rules that experts extract from normal images. VLMs adapted to inspection~\citep{xu2025customizing,gu2024anomalygpt} detect and explain defects, and SCAN~\citep{li2026scan} fine-tunes a multimodal LLM to verify written inspection criteria rule by rule. In these methods, the language model participates in detection at test time or through the code it generates. In our framework, the language model contributes to generating the synthetic anomalies, which are used only in training, and the anomaly score is computed by the classifier alone.

\subsection{Synthetic Data for Anomaly Detection}
\label{sec:related_synthetic}

Like many detectors, our classifier is trained on synthetic anomalies, which, for logical anomaly detection, must violate a rule of the category rather than merely alter local appearance. CutPaste~\citep{li2021cutpaste} pastes image patches and DRAEM~\citep{zavrtanik2021draem} blends external textures into normal images; such local changes suit structural defects but rarely break a logical rule. For structural defects, generate-then-train pipelines exist, in which an LLM or a VLM describes defects and a generative model adds them to images: AnomalyAny~\citep{sun2025anomalyanything} guides Stable Diffusion~\citep{rombach2022stablediffusion} with a normal image, AnomalyVFM~\citep{fucka2026anomalyvfm} inpaints defects into generated images to train a zero-shot detector, and MIRAGE~\citep{hu2026mirage} edits normal images to train a segmentation network without real anomalies. None of these pipelines is designed for logical constraints.

For logical anomalies, synthesis without real anomalies has used hand-designed rules: CSAD pastes extra components into training images to train its segmentation, SALAD erases or inpaints components in composition maps, and LogicAL~\citep{zhao2024logical} edits edge maps. ComGEN~\citep{tong2025comgen} instead learns a token for each component of an MVTec LOCO category with a diffusion model and removes tokens or samples unusual combinations to create anomalies; its detector compares each test image with the nearest normal image. All of these methods compare a test image with normal data at test time, directly or, in LogicAL, through a reconstruction of the image's normal appearance. What sets our framework apart is therefore not the use of synthetic anomalies but its test-time decision. Table~\ref{tab:synth_comparison} compares these approaches with our framework.

Beyond anomaly detection, generated training data usually keeps its label: DGInStyle~\citep{jia2024dginstyle} renders existing label maps in new styles, and ALIA~\citep{dunlap2023alia} edits training images along domain descriptions that an LLM summarizes from captions. Our edits instead change the label on purpose, turning a normal image into a specific rule violation, much as counterfactually augmented data~\citep{kaushik2020cad} revises text to fit a counterfactual label while avoiding unnecessary changes. Our image editor follows written instructions, like InstructPix2Pix~\citep{brooks2023instructpix2pix}. Because each unedited image is also labeled normal, the classifier can learn the difference between normal and violation on the same scene and largely under the same capture conditions. Out-of-distribution detection also creates outliers without real ones, but not by editing a given image: Dream-OOD~\citep{du2023dreamood} decodes outliers sampled from low-likelihood regions of a latent space into images, and EOE~\citep{cao2024eoe} has an LLM write the names of likely outlier classes.

\begin{table}[t]
    \centering
    \footnotesize
    \caption{\textbf{Synthetic anomalies in related methods and in our framework.} Rows follow a synthetic anomaly from its source to the decision at test time. A normal-reference score depends, at least in part, on a comparison with normal training data. Our framework differs from the first column in the anomalies it targets and from the other two in its test-time decision. Columns, from left: AnomalyAny, AnomalyVFM and MIRAGE~\citep{sun2025anomalyanything,fucka2026anomalyvfm,hu2026mirage}; ComGEN~\citep{tong2025comgen}; CSAD and SALAD~\citep{hsieh2024csad,fucka2025salad}.}
    \label{tab:synth_comparison}
    \setlength{\tabcolsep}{4.5pt}
    \renewcommand{\arraystretch}{1.5}
    \setlength{\aboverulesep}{0pt}
    \setlength{\belowrulesep}{0pt}
    \setlength{\extrarowheight}{2pt}
    \begin{tabular}{@{}>{\raggedright\arraybackslash}p{0.125\linewidth}*{3}{>{\raggedright\arraybackslash}p{0.17\linewidth}}>{\raggedright\arraybackslash}p{0.2\linewidth}}
        \toprule
         & AnomalyAny, AnomalyVFM, MIRAGE & ComGEN & CSAD, SALAD & \textbf{Ours} \\
        \midrule
        Target anomalies & Structural & Logical and structural & Mainly logical & Logical \\
        Knowledge source & Defect lists from an LLM or VLM & Component tokens learned from normal images & Components segmented in normal images & Written description of the category \\
        Generation & Diffusion generation, inpainting or editing & Token removal and low-density sampling in a diffusion model & Hand-designed edits of components or maps & Instruction-based editing of normal images into anomaly types \\
        Test-time decision & Detector trained on the synthetic data & Segmentation from differences to the nearest normal image & Fused scores, partly normal-reference & Classifier trained on normal images and synthetic anomalies; scores a test image without comparing it with normal data \\
        \bottomrule
    \end{tabular}
\end{table}

\subsection{Robustness in Anomaly Detection}
\label{sec:related_robustness}

Logical checks rest on counts, positions and proportions of parts; brightness reduction and zoom-out change how these appear: parts lose contrast against the background, or shrink and move. Yet robustness to capture shifts has been studied mainly for structural defects. Following corruption benchmarks for image classification~\citep{hendrycks2019benchmarking}, \citet{cao2023anomaly} corrupt MVTec AD test images, while AeBAD~\citep{zhang2023aebad}, RAD~\citep{cheng2024rad}, RobustAD~\citep{pemula2025robustad} and MVTec AD 2~\citep{heckler2025mvtecad2} capture real changes of illumination, viewpoint or background. Moreover, detectors that are almost perfect on MVTec AD degrade on real production data~\citep{baitieva2025beyond}. RobustAD also cautions that robustness to synthetic shifts may not carry over to real ones, and that training with the perturbations of a benchmark can leak its test distribution into training. We therefore make no claim about real capture shifts, and we report a variant of our framework trained without test-overlapping augmentations. We are not aware of a prior evaluation of MVTec LOCO logical anomaly detectors under capture shifts.

Test-time adaptation is an orthogonal route to robustness: Adapted-MoE~\citep{lei2025adaptedmoe} uses the test images themselves to correct the shift between the features of training and test samples (Section~\ref{sec:why_robust}).
\section{Proposed Method}
\label{sec:method}

\begin{figure}[H]
\centering
\resizebox{\textwidth}{!}{%
\begin{tikzpicture}[
    every node/.style={font=\footnotesize},
    head/.style={rectangle, draw=black, thick, align=center, inner sep=2pt, minimum height=1.45cm, execute at begin node={\hyphenpenalty=10000}},
    ex/.style={rectangle, draw=black, thick, align=left, inner sep=3pt, fill=white, minimum height=2.9cm, font=\scriptsize, execute at begin node={\hyphenpenalty=10000}},
    human/.style={fill=orange!25, dashed},
    model/.style={fill=gray!22},
    plain/.style={fill=white},
    ours/.style={fill=blue!12, line width=1.2pt},
    result/.style={fill=green!20},
    arr/.style={-{Stealth[length=2mm]}, thick},
    flow/.style={font=\scriptsize\itshape, text=black!80, align=center, inner sep=1pt},
    lane/.style={rounded corners=4pt, draw=gray!55, fill=gray!6},
    lanetitle/.style={font=\footnotesize\bfseries, text=black!70, anchor=west}
]
\draw[lane] (-0.05, 0.55) rectangle (13.45, -4.5);
\draw[lane] (-0.05, -5.3) rectangle (13.45, -7.95);
\node[lanetitle] at (0.0, 0.28) {Anomaly design (example: Pushpins)};
\node[lanetitle, anchor=east] at (13.4, -5.55) {Training and test};

\node[head, human, text width=2.36cm, anchor=north] (h1) at (1.3, 0)
    {\textbf{Category}\\\textbf{description}};
\node[head, model, text width=2.96cm, anchor=north] (h2) at (5.1, 0)
    {\textbf{Language model}\\proposes $N$ anomaly types and an editing prompt for each};
\node[head, model, text width=2.06cm, anchor=north] (h3) at (8.75, 0)
    {\textbf{Image editor}\\applies a prompt to a normal image};
\node[head, human, text width=2.36cm, anchor=north] (h4) at (12.1, 0)
    {\textbf{Curation}\\VLM-assisted};

\node[ex, dashed, text width=2.29cm, anchor=north] (e1) at (h1.south)
    {\textit{Each of the 15 compartments holds exactly one pushpin.}};
\node[ex, text width=2.89cm, anchor=north] (e2) at (h2.south)
    {\textbf{Type 6:} One compartment contains two pushpins.\\[4pt]
     \textbf{Prompt:} Add an extra pushpin to one compartment so it contains two pushpins.};
\node[ex, text width=1.99cm, anchor=north] (e3) at (h3.south) {};
\node[ex, dashed, text width=2.29cm, anchor=north] (e4) at (h4.south)
    {\textbf{Kept} as a synthetic anomaly of type 6.\\[4pt]
     Edits that show no valid anomaly are discarded.};

\node[inner sep=0, anchor=north] (gimg) at ([yshift=-0.12cm]e3.north)
    {\includegraphics[width=1.95cm]{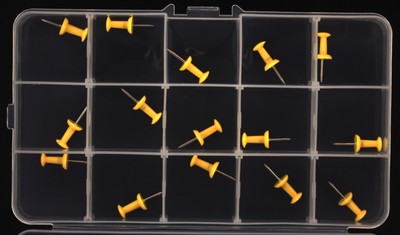}};
\node[inner sep=0, anchor=south] (simg) at ([yshift=0.12cm]e3.south)
    {\includegraphics[width=1.95cm]{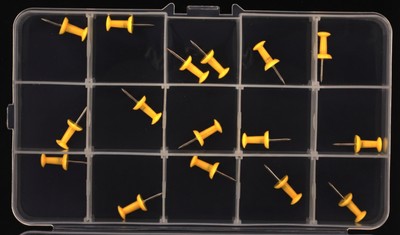}};
\draw[-{Stealth[length=1.5mm]}, semithick] (gimg.south) -- (simg.north);
\draw[red, thick] ($(simg.south west)+(0.99,0.91)$) ellipse (0.25cm and 0.22cm);

\draw[arr] (e1.east) -- node[flow, above=2pt] {text} (e2.west);
\draw[arr] (e2.east) -- node[flow, above=2pt] {type +\\prompt} (e3.west);
\draw[arr] (e3.east) -- node[flow, above=2pt] {edited\\image} (e4.west);

\node[head, plain, text width=2.16cm, anchor=north, minimum height=1.95cm] (good) at (1.2, -5.8)
    {\textbf{Normal images}\\train and\\validation splits};
\node[head, plain, text width=2.76cm, anchor=north, minimum height=1.95cm] (sub) at (4.8, -5.8)
    {\textbf{Normal sub-mode}\\\textbf{discovery}\\clusters pretrained features; $K{=}1$ if no clear groups};
\node[head, ours, text width=2.46cm, anchor=north, minimum height=1.95cm] (clf) at (8.6, -5.8)
    {\textbf{Classifier}\\ConvNeXt-L (DINOv3);\\$K$ normal classes vs.\ $N_c$ anomaly types};
\node[head, result, text width=2.31cm, anchor=north, minimum height=1.95cm] (test) at (12.125, -5.8)
    {\textbf{Score $s(x)$}\\Eq.~(\ref{eq:anomaly_score}), for a test image~$x$; no normal reference};

\draw[arr] (good) -- node[flow, above=2pt] {normal\\images} (sub);
\draw[arr] (sub) -- node[flow, above=2pt] {$K$ normal\\classes} (clf);
\draw[arr] (clf) -- node[flow, above=2pt] {softmax} (test);

\draw[arr] (good.north) -- (1.2, -4.9) -| node[flow, pos=0.25, above=1pt] {normal images to edit, one prompt each} ([xshift=-0.45cm]e3.south);
\draw[arr] (e4.south) -- (12.1, -4.9) -| node[flow, pos=0.13, below=1pt] {synthetic anomalies, labeled by type} ([xshift=0.45cm]clf.north);
\end{tikzpicture}}%
\caption{\textbf{Overview of our framework, with an example from Pushpins.} Top: a person describes the category; the language model proposes anomaly types that break the description and writes an editing prompt for each; the image editor applies a prompt to a normal image; curation keeps the valid edits (red ellipse: two pushpins in one compartment). Bottom: the classifier, trained on $K$ normal classes and the anomaly types, scores a test image on its own. Arrows show what passes between steps. Dashed orange boxes involve a person; gray boxes are pretrained models. The blue box is the only component used at test time. The edited image is generated by the image editor (Gemini 3 Pro Image), a generative AI model.}
\label{fig:pipeline}
\end{figure}
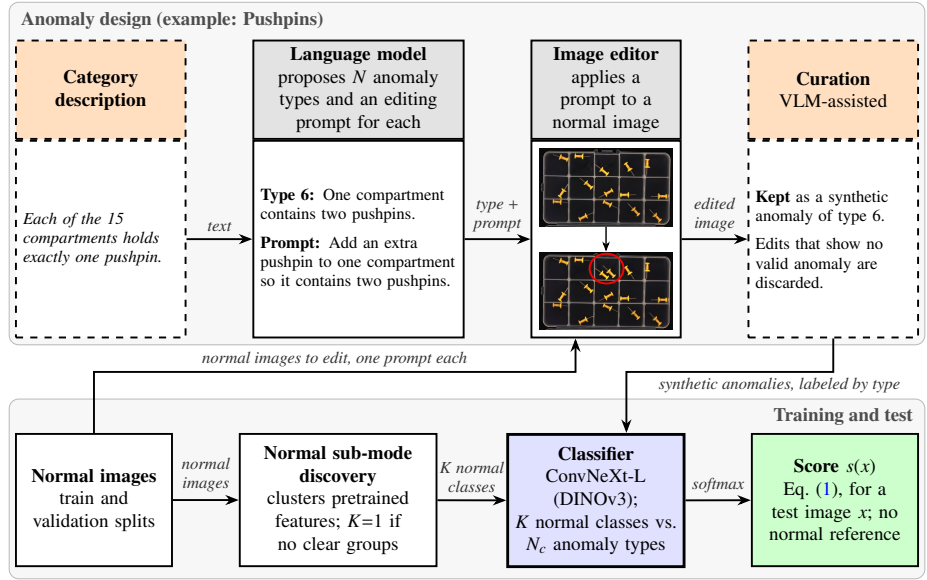

Fig.~\ref{fig:pipeline} gives an overview of our framework with an example from Pushpins. Section~\ref{sec:problem} formulates the problem. Section~\ref{sec:generation} describes how synthetic anomalies are generated from a written description of each category, and Section~\ref{sec:submodes} how the normal images are divided into sub-modes. Section~\ref{sec:classification} presents the classifier and its training, Section~\ref{sec:augmentation} the training augmentation, and Section~\ref{sec:evaluation} the evaluation protocol under capture shifts.

\subsection{Problem Formulation}
\label{sec:problem}

The MVTec LOCO benchmark~\citep{bergmann2022beyond} provides training and validation splits with only normal (good) images and a test split with normal images, logical anomalies and structural anomalies. We restrict the evaluation to the normal test images and the logical anomalies, which violate constraints on the presence, number, position or combination of parts, and report image-level AUROC. Our framework adds synthetic anomalies to the training and validation splits by editing their normal images. The classifier learns $K$ normal classes (Section~\ref{sec:submodes}) and $N_c$ anomaly types (Section~\ref{sec:generation}) and is selected on the validation split, so no real anomalous image is used for training or checkpoint selection. At test time, the classifier scores an image alone, without comparing it with normal data. We call this way of scoring a fully discriminative decision.

\subsection{Anomaly Design and Generation}
\label{sec:generation}

The classifier needs examples of rule violations, but the training split holds only normal images, so we create such violations from a description of the product.

\noindent\textbf{Category description.}
A person writes a free-text description of the normal images: which parts a correct product contains, how many, where they are and which rules relate them. For Pushpins, for example, each of the 15 compartments holds exactly one pushpin. Human knowledge of the product is a common input in logical anomaly detection: LogSAD~\citep{zhang2025logsad} builds its matching rules from GPT-4V proposals prompted with the constraint descriptions of the MVTec LOCO paper, LogicQA~\citep{kwon2025logicqa} gives the same descriptions to a VLM to write checklist questions, LogicAD~\citep{jin2025logicad} turns a written description of normality into logical formulas, LogiCode~\citep{zhang2025logicode} turns rules that experts extract from normal images into code, and SCAN~\citep{li2026scan} checks written inspection criteria rule by rule. \ref{app:descriptions} gives our descriptions in the structured form passed to the language model.

\noindent\textbf{Anomaly types and editing prompts.}
From the category description and a fixed definition of a logical anomaly, the language model (GPT-5; \citealp{openai2025gpt5}) proposes anomaly types, each a short statement of one rule violation, and writes one editing prompt per type. We set the number of types to $N=20$ per category, which is within the range of prior work, from 10 defect types per category in MIRAGE~\citep{hu2026mirage} to 50 logical-anomaly descriptions requested in LogSAD's GPT-4V prompt~\citep{zhang2025logsad}. For Pushpins, type 6 is ``One compartment contains two pushpins'', with the prompt ``Add an extra pushpin to one compartment so it contains two pushpins'' (Fig.~\ref{fig:pipeline}). A person then checked the types and prompts. \ref{app:generation} gives all instructions, and \ref{app:types} all types and prompts.

\noindent\textbf{Image editing.}
The image editor, Gemini 3 Pro Image~\citep{gemini3proimage}, applies one prompt to each normal image of the training and validation splits, and each resulting synthetic anomaly takes the anomaly type of its prompt as its label. Every normal image yields one synthetic anomaly, so $N$ sets only how the synthetic anomalies are divided among the types. Synthetic validation anomalies are made from validation images, so they share no source image with the training set.

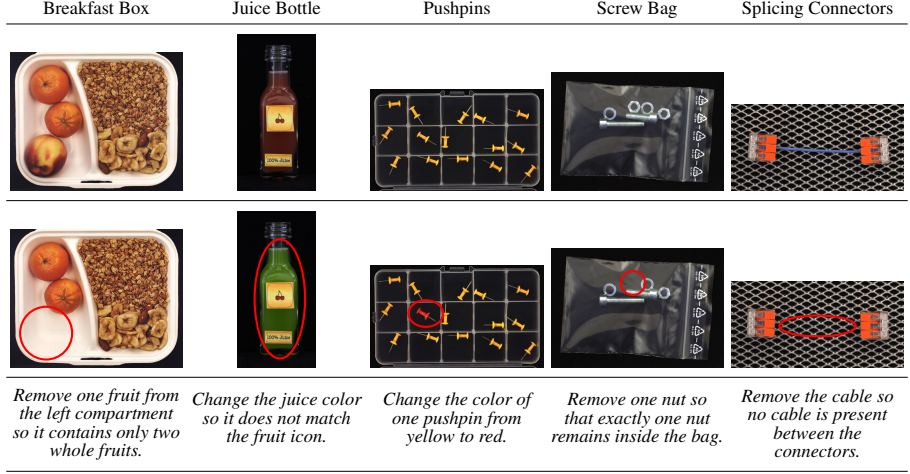
\begin{figure}[H]
    \centering
    \newlength{\cellw}\setlength{\cellw}{0.188\textwidth}%
    \newlength{\cellh}\setlength{\cellh}{2.2cm}%
    \setlength{\tabcolsep}{1.5pt}%
    \setlength{\arrayrulewidth}{0.3pt}%
    \newcommand{\markedimg}[5]{\begin{tikzpicture}
        \node[inner sep=0, anchor=south west] (im) at (0,0) {\includegraphics[width=\cellw,height=\dimexpr\cellh-4pt\relax,keepaspectratio]{#1}};
        \begin{scope}[x={(im.south east)}, y={(im.north west)}]
            \draw[red, thick] (#2,#3) ellipse [x radius=#4, y radius=#5];
        \end{scope}\end{tikzpicture}}%
    \begin{tabular}{>{\centering\arraybackslash}m{\cellw}>{\centering\arraybackslash}m{\cellw}>{\centering\arraybackslash}m{\cellw}>{\centering\arraybackslash}m{\cellw}>{\centering\arraybackslash}m{\cellw}}
        \multicolumn{1}{c}{\scriptsize Breakfast Box} & \multicolumn{1}{c}{\scriptsize Juice Bottle} & \multicolumn{1}{c}{\scriptsize Pushpins} & \multicolumn{1}{c}{\scriptsize Screw Bag} & \multicolumn{1}{c}{\scriptsize Splicing Connectors} \\
        \hline
        \rule{0pt}{\cellh}\includegraphics[width=\cellw,height=\dimexpr\cellh-4pt\relax,keepaspectratio]{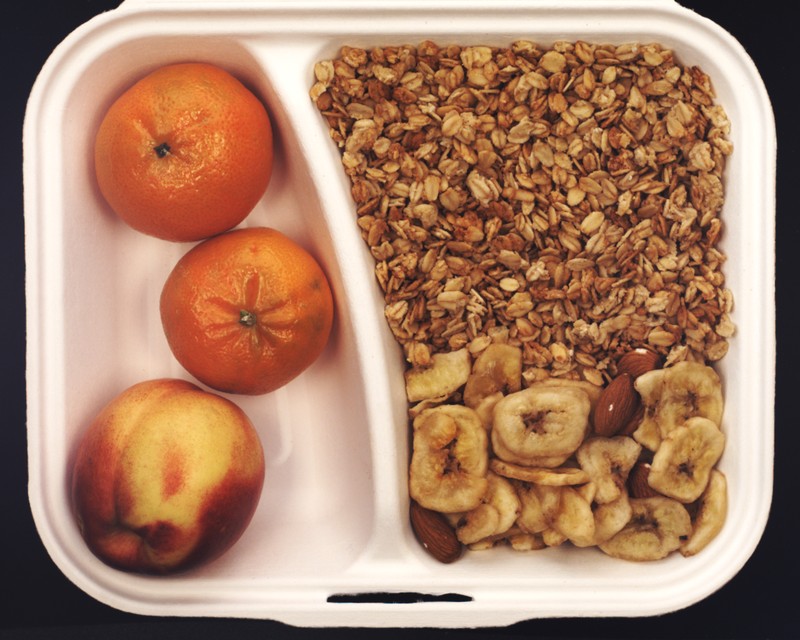} &
        \rule{0pt}{\cellh}\includegraphics[width=\cellw,height=\dimexpr\cellh-4pt\relax,keepaspectratio]{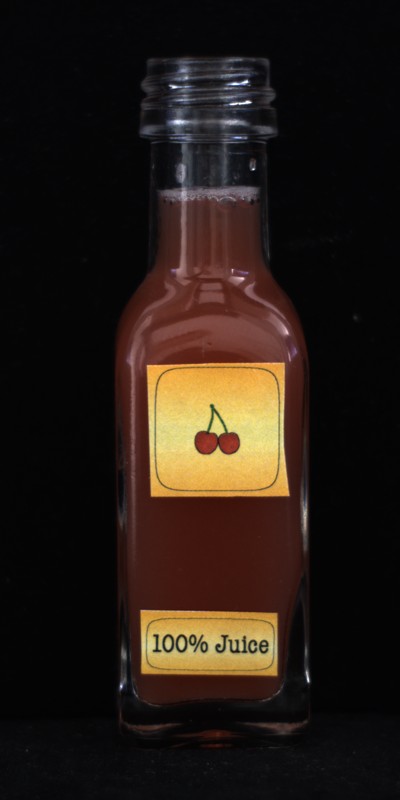} &
        \rule{0pt}{\cellh}\includegraphics[width=\cellw,height=\dimexpr\cellh-4pt\relax,keepaspectratio]{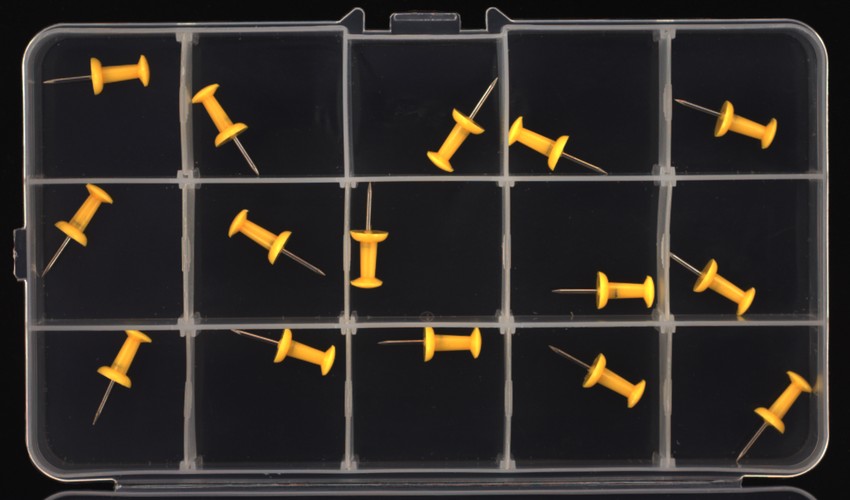} &
        \rule{0pt}{\cellh}\includegraphics[width=\cellw,height=\dimexpr\cellh-4pt\relax,keepaspectratio]{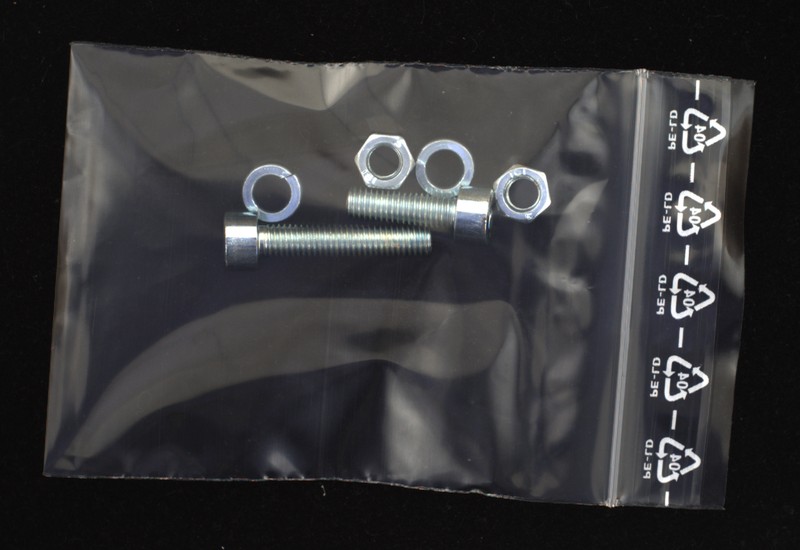} &
        \rule{0pt}{\cellh}\includegraphics[width=\cellw,height=\dimexpr\cellh-4pt\relax,keepaspectratio]{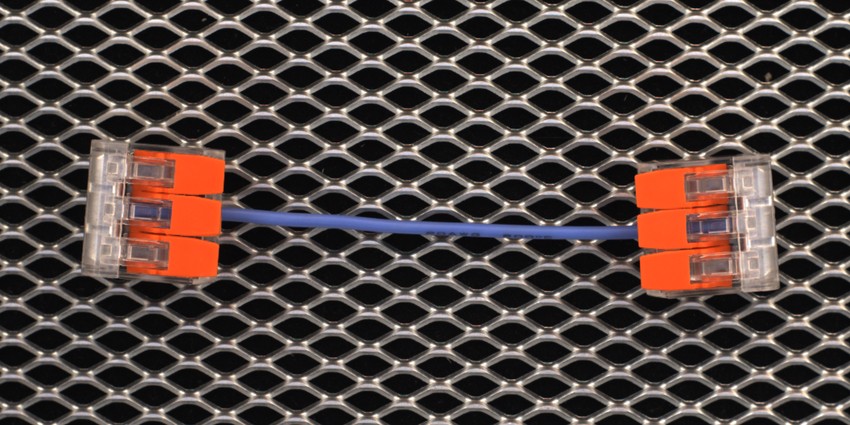} \\
        \hline
        \rule{0pt}{\cellh}\markedimg{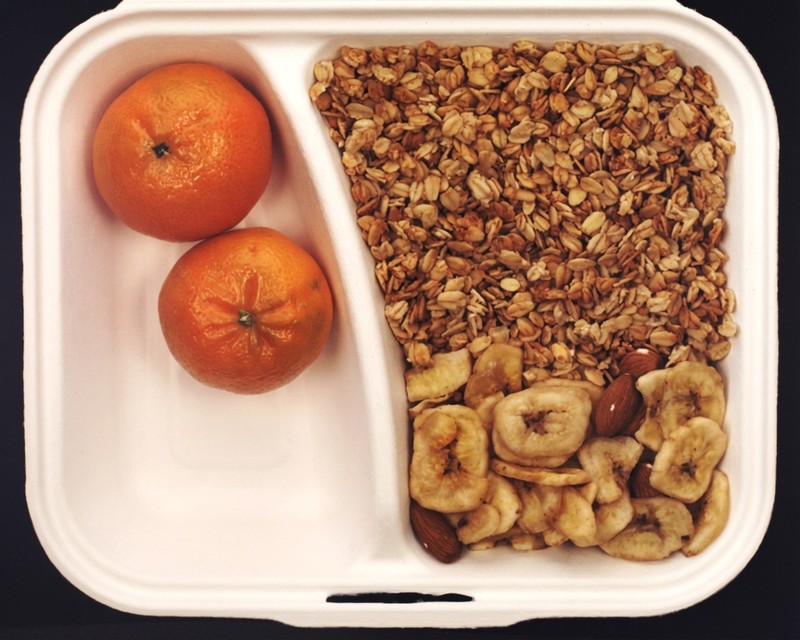}{0.210}{0.250}{0.150}{0.190} &
        \rule{0pt}{\cellh}\markedimg{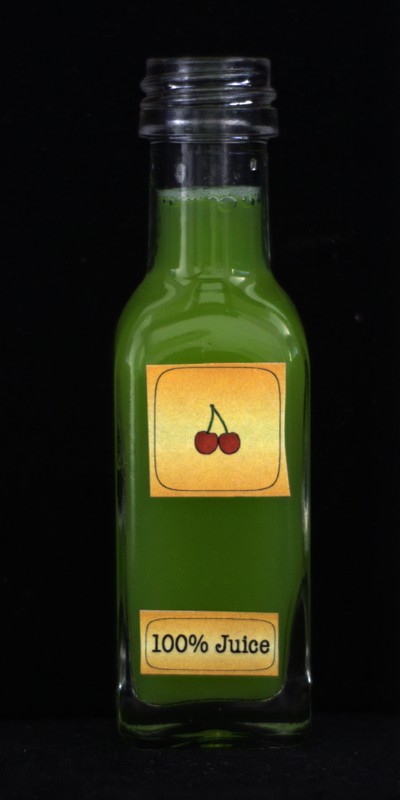}{0.540}{0.440}{0.320}{0.380} &
        \rule{0pt}{\cellh}\markedimg{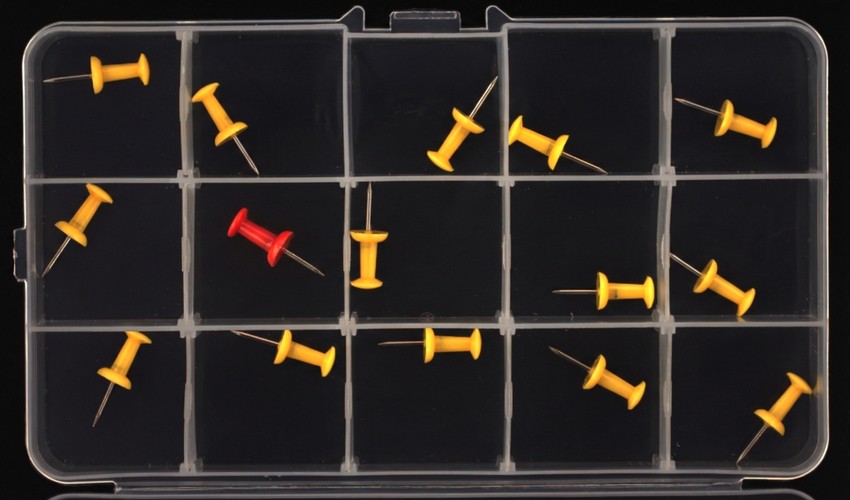}{0.320}{0.530}{0.100}{0.130} &
        \rule{0pt}{\cellh}\markedimg{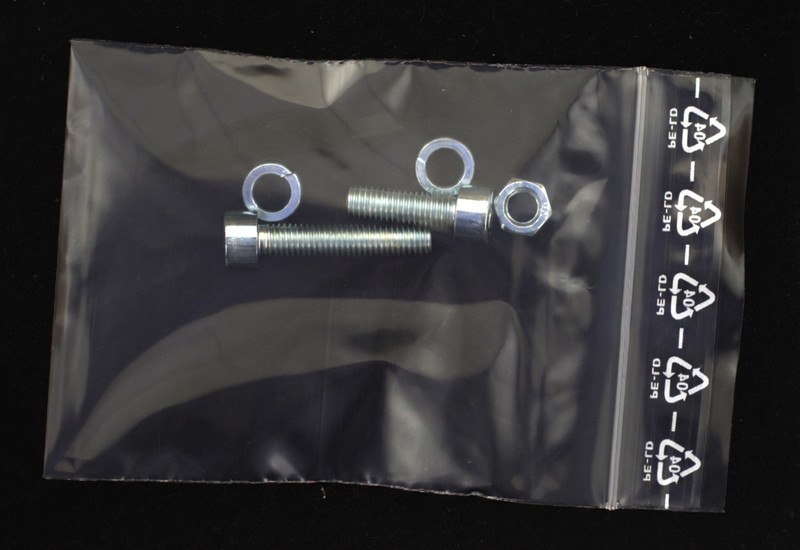}{0.475}{0.710}{0.070}{0.100} &
        \rule{0pt}{\cellh}\markedimg{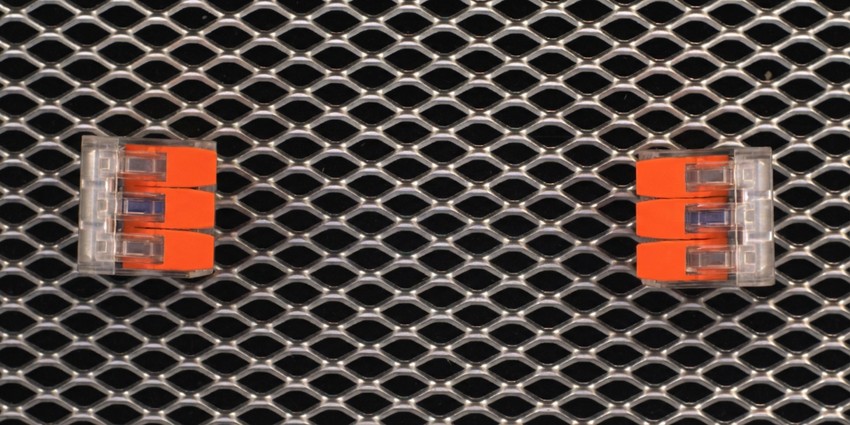}{0.500}{0.480}{0.220}{0.120} \\
        \hline
        \parbox[c]{\cellw}{\vspace{4pt}\centering\scriptsize\baselineskip=7pt\textit{Remove one fruit from the left compartment so it contains only two whole fruits.}\vspace{4pt}} &
        \parbox[c]{\cellw}{\vspace{4pt}\centering\scriptsize\baselineskip=7pt\textit{Change the juice color so it does not match the fruit icon.}\vspace{4pt}} &
        \parbox[c]{\cellw}{\vspace{4pt}\centering\scriptsize\baselineskip=7pt\textit{Change the color of one pushpin from yellow to red.}\vspace{4pt}} &
        \parbox[c]{\cellw}{\vspace{4pt}\centering\scriptsize\baselineskip=7pt\textit{Remove one nut so that exactly one nut remains inside the bag.}\vspace{4pt}} &
        \parbox[c]{\cellw}{\vspace{4pt}\centering\scriptsize\baselineskip=7pt\textit{Remove the cable so no cable is present between the connectors.}\vspace{4pt}} \\
        \hline
    \end{tabular}
    \caption{\textbf{Normal images and synthetic anomalies.} Each column shows a normal training image, the synthetic anomaly made from it (red ellipse: edited region) and its editing prompt. All five synthetic anomalies passed curation and are in the final training data. The synthetic anomalies are generated by the image editor (Gemini 3 Pro Image), a generative AI model.}
    \label{fig:examples}
\end{figure}

\noindent\textbf{Curation.}
An edit does not always produce a valid anomaly. We therefore inspected the edited images with the help of a VLM and human verification, and kept only valid logical anomalies. Anomaly types with too few valid images are not used as classes (\ref{app:data}); $N_c$ denotes the number of types that remain. Table~\ref{tab:data} lists the resulting data, and Fig.~\ref{fig:examples} shows examples.

\subsection{Discovering Sub-Modes of Normal Images}
\label{sec:submodes}

The normal images of a category are not always homogeneous and may form several distinct sub-modes. The normal training images of Juice Bottle, for instance, show three distinct products: cherry, orange and banana juice, each with its matching fruit icon. If these products share a single normal class, the classifier may learn each icon and each juice color as normal in isolation and fail to detect an incorrect combination, which is precisely a logical anomaly. Each sub-mode is therefore assigned its own normal class.

The sub-modes are identified automatically from the normal training images alone. Features are extracted with the pretrained backbone of the classifier (ConvNeXt-L with DINOv3 weights) before any fine-tuning and clustered with a Gaussian mixture model. The number of clusters $K$ is proposed by cross-validated likelihood, and the resulting partition is accepted only if the clusters are well separated, none of them is small and they are stable under resampling; otherwise, the next smaller $K$ is tested, down to a single normal class. Each normal training image takes its cluster as a pseudo-label, fixed before training, and the classifier is then trained with these labels and the anomaly types by standard cross-entropy. Each validation image is assigned the pseudo-label of its most likely cluster. \ref{app:submodes} lists the thresholds.

The procedure splits Juice Bottle into $K=4$ classes, which follow the three fruit types with the banana images in two clusters, and Splicing Connectors into $K=3$, one per connector type; the other categories keep one normal class (Table~\ref{tab:submodes}). The two banana clusters are not a distinct product, but both are normal classes, so the extra split does not change which images count as normal. The idea itself is not new: splitting data into pseudo-classes or subclasses by clustering is well established~\citep{ionescu2019object,cohen2022ood,sohoni2020george,liao2026pgbl}, and LogSAD~\citep{zhang2025logsad} samples the normal images that it gives to GPT-4V from the known normal sub-classes of each MVTec LOCO category. We adapt the idea to logical anomaly detection with synthetic anomalies.

\subsection{Classification Model and Training Procedure}
\label{sec:classification}

\noindent\textbf{Classifier and anomaly score.}
The classifier is a ConvNeXt-L~\citep{liu2022convnext} with DINOv3 pretrained weights~\citep{simeoni2025dinov3} and a linear head with one output per normal class and per anomaly type (Table~\ref{tab:hyper}). It is fine-tuned end to end with cross-entropy on the normal images, labeled with their pseudo-labels, and on the synthetic anomalies, labeled by anomaly type. The anomaly score, Eq.~(\ref{eq:anomaly_score}), is the probability mass outside the normal classes,
\begin{equation}
    s(x) = 1 - \sum_{k \in \mathcal{G}} \operatorname{softmax}\big(f(x)\big)_k ,
    \label{eq:anomaly_score}
\end{equation}
where $f(x)$ are the logits of the classifier and $\mathcal{G}$ holds the indices of the $K$ normal classes.

\noindent\textbf{Training.}
We fine-tune the classifier for 30 epochs at $448\times448$ pixels with AdamW~\citep{loshchilov2019adamw}; Table~\ref{tab:hyper} lists all settings. We keep the checkpoint with the lowest cross-entropy on the validation split, which holds the real validation normals with their pseudo-labels and the curated synthetic validation anomalies. This validation-loss checkpoint uses no test image or label. We do not select by validation AUROC, because it saturates at 100\% early in training and then no longer distinguishes between checkpoints.

\subsection{Training Augmentation}
\label{sec:augmentation}

Because some standard augmentation operations resemble the test perturbations, we train two variants of the classifier. The first variant, the full recipe, passes normal images and synthetic anomalies through the same random augmentation pipeline~\citep{buslaev2020albumentations}: flips and $90^\circ$ rotations, zoom-out with edge-replicated borders, blur, noise and compression, a photometric block (brightness and contrast, contrast-limited adaptive histogram equalization (CLAHE), tone curve, sharpening), and emboss or posterize (Table~\ref{tab:augment}). Its zoom-out and brightness change cover the same ranges as the two test perturbations (Section~\ref{sec:evaluation}). The second variant, trained without test-overlapping augmentations, omits the zoom-out and photometric blocks and is otherwise identical; \citet{cao2023anomaly} likewise exclude augmentations that overlap their test corruptions. Since neither test perturbation occurs in its training data, our robustness conclusions are based on this variant. Generic augmentations such as blur and noise remain and may contribute to the robustness (Section~\ref{sec:why_robust}). The full recipe gives the higher clean AUROC, but its robustness could come from its overlap with the test perturbations.

\subsection{Evaluation Protocol}
\label{sec:evaluation}

\begin{figure}[H]
    \centering
    \begin{tikzpicture}
    \def\w{1.6cm}     %
    \def\s{1.76}      %
    \node[inner sep=0] (b0) at (0,0) {\includegraphics[width=\w]{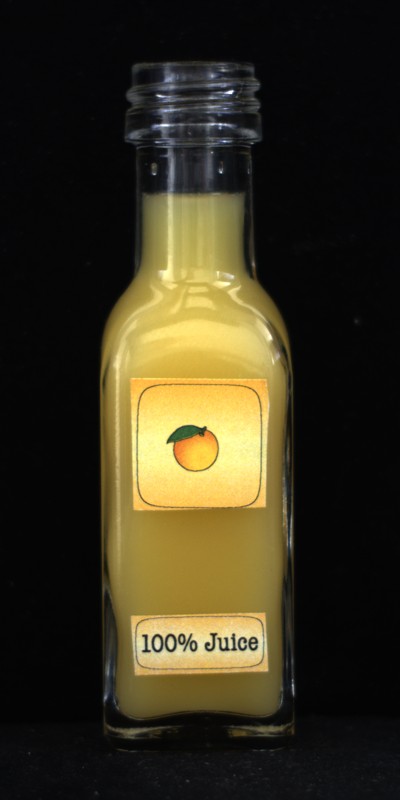}};
    \node[inner sep=0] (b1) at (\s,0) {\includegraphics[width=\w]{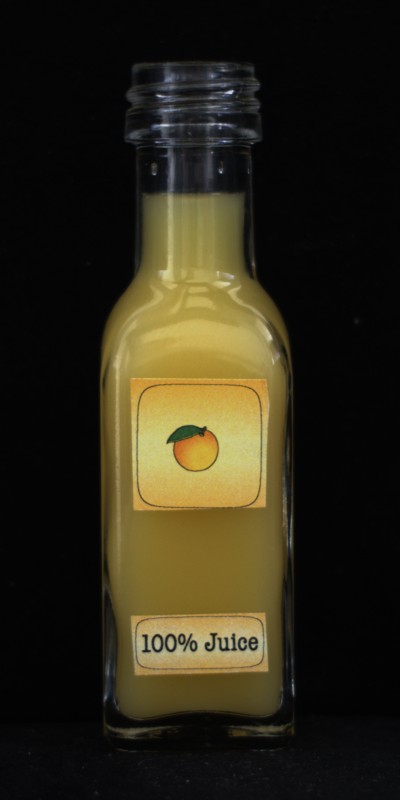}};
    \node[inner sep=0] (b2) at (2*\s,0) {\includegraphics[width=\w]{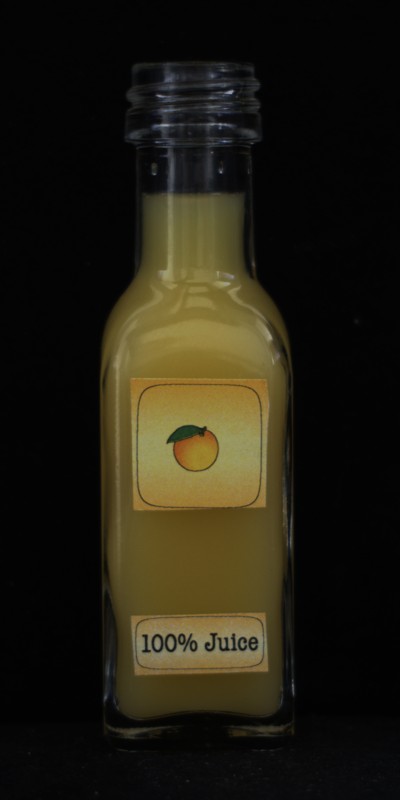}};
    \node[inner sep=0] (b3) at (3*\s,0) {\includegraphics[width=\w]{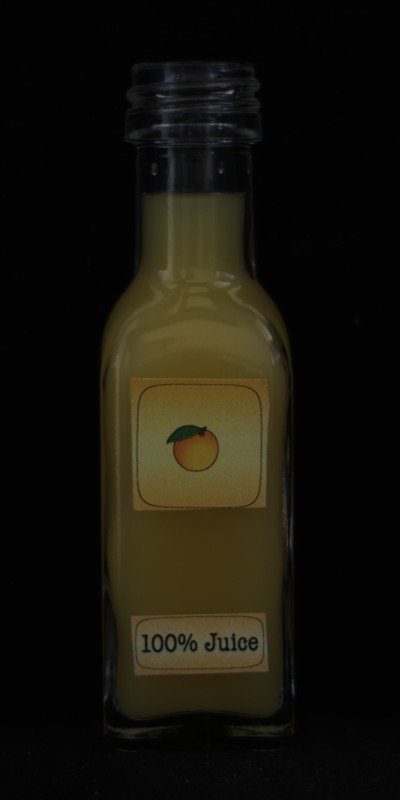}};
    \node[font=\scriptsize\rmfamily] at ([yshift=-2mm]b0.south) {Clean};
    \node[font=\scriptsize\rmfamily] at ([yshift=-2mm]b1.south) {20\%};
    \node[font=\scriptsize\rmfamily] at ([yshift=-2mm]b2.south) {40\%};
    \node[font=\scriptsize\rmfamily] at ([yshift=-2mm]b3.south) {60\%};
    \draw[-stealth, gray!60!black, semithick]
        ([yshift=-6mm]b0.south) -- ([yshift=-6mm]b3.south);
    \node[font=\scriptsize\rmfamily, fill=white, inner sep=1pt]
        at ([yshift=-6mm]$(b1.south)!0.5!(b2.south)$) {Brightness reduction};
    \node[inner sep=0] (s0) at (0,-4.8) {\includegraphics[width=\w]{000_clean.jpg}};
    \node[inner sep=0] (s1) at (\s,-4.8) {\includegraphics[width=\w]{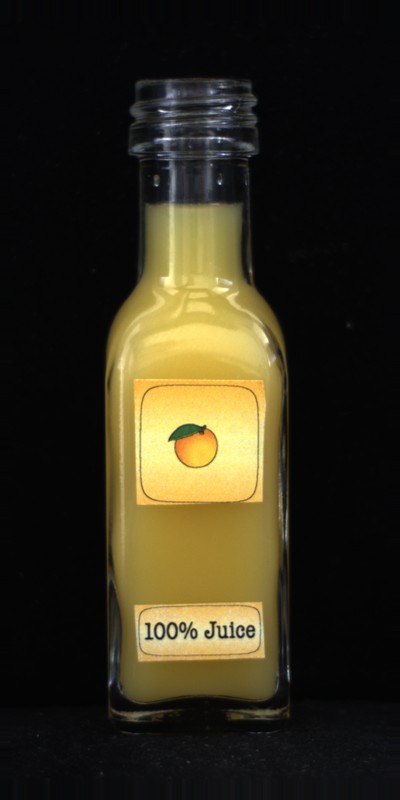}};
    \node[inner sep=0] (s2) at (2*\s,-4.8) {\includegraphics[width=\w]{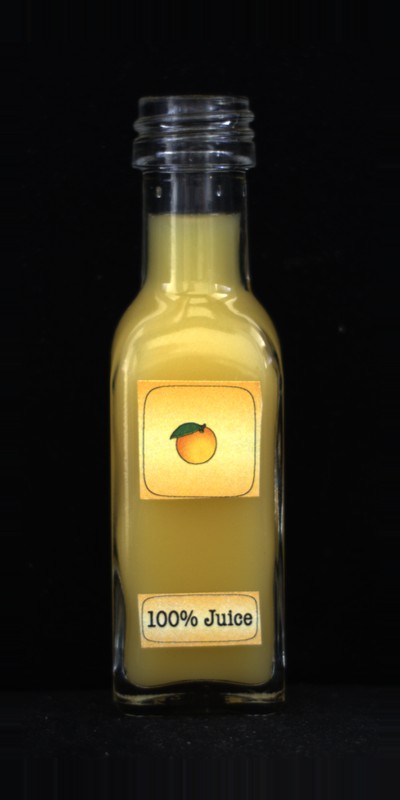}};
    \node[inner sep=0] (s3) at (3*\s,-4.8) {\includegraphics[width=\w]{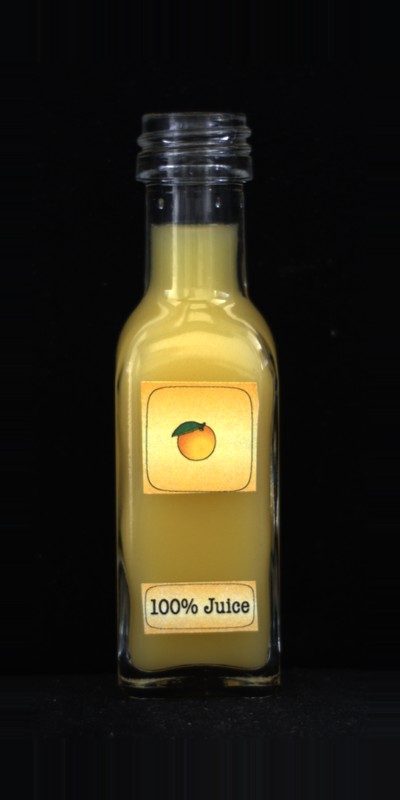}};
    \node[font=\scriptsize\rmfamily] at ([yshift=-2mm]s0.south) {Clean};
    \node[font=\scriptsize\rmfamily] at ([yshift=-2mm]s1.south) {5\%};
    \node[font=\scriptsize\rmfamily] at ([yshift=-2mm]s2.south) {10\%};
    \node[font=\scriptsize\rmfamily] at ([yshift=-2mm]s3.south) {15\%};
    \draw[-stealth, gray!60!black, semithick]
        ([yshift=-6mm]s0.south) -- ([yshift=-6mm]s3.south);
    \node[font=\scriptsize\rmfamily, fill=white, inner sep=1pt]
        at ([yshift=-6mm]$(s1.south)!0.5!(s2.south)$) {Zoom-out};
    \end{tikzpicture}
    \caption{\textbf{Perturbation examples (Juice Bottle).} Each row starts from the clean image. Top: brightness reduced by 20\%, 40\% and 60\%. Bottom: zoom-out by 5\%, 10\% and 15\% with edge-replicated borders.}
    \label{fig:perturbations}
\end{figure}

\noindent\textbf{Perturbations.}
Robustness is measured on test images with simulated capture shifts. We apply six perturbations from two families to the full-resolution test images, before any preprocessing of the evaluated method (Fig.~\ref{fig:perturbations}). Brightness reduction multiplies all pixel values by 0.8, 0.6 or 0.4. Zoom-out shrinks the image by 5\%, 10\% or 15\% and pads it back to its original size with edge-replicated borders, so every object stays in view. Together with the unperturbed (clean) condition, the six perturbations define seven test conditions, denoted Clean, B20, B40 and B60 for brightness reduction, and Z5, Z10 and Z15 for zoom-out.

We chose these two families because they are typical uncontrolled capture shifts on an inspection line, due to changes in illumination and camera placement~\citep{baitieva2025beyond}; because they act on the whole image without changing its logical content; and because their severity can be graded. We reduce rather than increase brightness because an increase clips bright regions to uniform white, which can erase parts and boundaries and thus alter the logical content of the image; a reduction preserves all parts, at the cost of coarser quantization in dark regions. Likewise, we zoom out rather than in, because zooming in would crop parts at the border and could itself create an anomaly. Other perturbation families and real capture shifts are left for future work.

\noindent\textbf{Heterogeneous perturbation protocol.}
A deployed detector usually compares every score with the same threshold, whatever the condition of the image. In the heterogeneous perturbation protocol, each test image therefore independently receives one of the seven conditions with equal probability, and one AUROC is computed over the mixed test set; \citet{baitieva2025beyond} applied a similar random drift, drawn per test image, to real industrial datasets, and the mixed test split of MVTec AD 2, a benchmark of mainly structural defects, likewise draws the lighting of each test image at random~\citep{heckler2025mvtecad2}. We repeat the random assignment 100 times, use the same assignments for every method, and report the mean as the heterogeneous AUROC.

\noindent\textbf{Metrics.}
We use the clean AUROC $A_{\mathrm{clean}}$, the heterogeneous AUROC $A_{\mathrm{het}}$, and the per-condition AUROC $A_{\mathrm{pc}}$, for which each perturbed condition is scored on its own and the six AUROCs are averaged. We report the drop and the relative drop of Eq.~(\ref{eq:relative_loss}),
\begin{equation}
    \text{drop} = A_{\mathrm{clean}} - A_{\mathrm{het}}, \qquad
    \text{relative drop} = 100 \times \frac{A_{\mathrm{clean}} - A_{\mathrm{het}}}{A_{\mathrm{clean}} - 50},
    \label{eq:relative_loss}
\end{equation}
where 50 is the AUROC of chance; both are positive when AUROC falls, and the relative drop is the percentage of the clean AUROC above chance that is lost. The drop splits into two parts. The within-condition drop, $A_{\mathrm{clean}} - A_{\mathrm{pc}}$, occurs even when all test images share one condition: the method separates anomalies from normal images less well within a darkened or zoomed-out test set. The cross-condition drop, $A_{\mathrm{pc}} - A_{\mathrm{het}}$, arises because the conditions are mixed and the scores of different conditions are compared with each other. The worst condition is the perturbed condition with the lowest single-condition AUROC.
\section{Experimental Evaluation}
\label{sec:experiments}

\subsection{Experimental Setup}
\label{sec:setup}

\paragraph{Dataset}
We use all five categories (Breakfast Box, Juice Bottle, Pushpins, Screw Bag and Splicing Connectors) of MVTec LOCO AD~\citep{bergmann2022beyond} with the standard splits. Training and validation use real normal (good) images and synthetic anomalies. The test set contains only real images: normal images and logical anomalies; we leave the structural anomalies out. Table~\ref{tab:data} gives the image counts.

\paragraph{Implementation}
The classifier follows the recipe and checkpoint selection of Section~\ref{sec:classification}, in two versions (Section~\ref{sec:augmentation}): Ours (no overlap), without test-overlapping augmentations (no zoom-out or brightness augmentation in training), and Ours (full recipe).

\paragraph{Baselines}
We compare with four normal-reference methods, CSAD~\citep{hsieh2024csad}, SALAD~\citep{fucka2025salad}, LogSAD~\citep{zhang2025logsad} and PatchCore~\citep{roth2022patchcore}, and with AnomalyCLIP~\citep{zhou2024anomalyclip}, a zero-shot method that uses no normal reference. All baselines use the official code and released weights without training any network, and each keeps the same model for all seven test conditions; for CSAD, SALAD and LogSAD, our clean measurements are within one point of the reported values. SALAD scores a composition map of each test image; because the trained model that produces these maps is not released, we use the maps that SALAD's authors released for the clean test images, so these maps are unaffected by the perturbation, which favors SALAD. \ref{app:baselines} gives the settings.

\paragraph{Metrics}
We report image-level AUROC on the logical anomalies, in percent and averaged over the five categories, together with the drop and the relative drop (Section~\ref{sec:evaluation}).

\subsection{Main Results}
\label{sec:main_results}

\begin{table}[t]
\centering
\footnotesize\setlength{\tabcolsep}{4pt}
\caption{\textbf{Main results.} Clean and heterogeneous image-level logical AUROC (\%), mean over the five categories.}
\label{tab:main}
\begin{tabular*}{\textwidth}{@{\extracolsep{\fill}}>{\raggedright\arraybackslash}p{4.6cm}ccccc@{}}
\toprule
& & & Hetero- & Drop & Relative \\
Method & Reported & Clean & geneous & (points) & drop (\%) \\
\midrule
\textbf{Ours (no overlap)} & -- & 93.2 & 92.5 & 0.7 & 1.6 \\
\textbf{Ours (full recipe)} & -- & 94.9 & 94.1 & 0.8 & 1.8 \\
\midrule
CSAD & 96.7 & 97.7 & 68.1 & 29.6 & 62.1 \\
SALAD$^{a}$ & 96.5 & 96.4 & 62.3 & 34.1 & 73.5 \\
LogSAD & 89.3 & 89.2 & 73.9 & 15.3 & 39.0 \\
PatchCore$^{b}$ & -- & 75.4 & 65.7 & 9.7 & 38.2 \\
\midrule
AnomalyCLIP$^{c}$ & -- & 57.3 & 54.3 & 3.0 & 41.1 \\
\midrule
SALAD$^\dagger$ & 98.7 & -- & -- & -- & -- \\
PSAD & 98.1 & -- & -- & -- & -- \\
SAM-LAD & 95.8 & -- & -- & -- & -- \\
\bottomrule
\end{tabular*}
\par\smallskip\begin{minipage}{\textwidth}\footnotesize
Reported: five-category logical AUROC from the method's own paper; the other columns are our measurements, and our clean values for CSAD, SALAD and LogSAD are within one point of the reported ones. Drop = clean $-$ heterogeneous, in points; relative drop = $100\times\text{drop}/(\text{clean}-50)$, the share of the above-chance clean AUROC that is lost; both are positive when AUROC falls. Ours (no overlap): trained without test-overlapping augmentations. Bottom block: methods with only a reported clean result, whose robustness we could not measure. \ref{app:baselines} gives the settings of all baselines. $^{a}$~We run SALAD with its own unsupervised composition maps; SALAD$^\dagger$ uses PSAD's supervised part maps. The trained composition-map model is not released, so we use the maps that SALAD's authors released for the clean images; these maps are therefore unaffected by the perturbation, which favors SALAD. $^{b}$~PatchCore reports no MVTec LOCO result. $^{c}$~AnomalyCLIP uses no normal reference and is near chance on clean images.
\end{minipage}
\end{table}

We first quantify the performance degradation of existing methods under capture shifts (Table~\ref{tab:main}, Fig.~\ref{fig:teaser}). Under the heterogeneous perturbation protocol, the normal-reference methods we tested lose 9.7 to 34.1 points, a relative drop of 38.2\% to 73.5\%. Notably, among these methods, the two with the highest clean AUROC exhibit the largest degradation: CSAD drops from 97.7\% to 68.1\% and SALAD from 96.4\% to 62.3\%. AnomalyCLIP, near chance, loses only 3.0 points, which is still a relative drop of 41.1\%. AnomalyCLIP's low clean AUROC agrees with a related zero-shot evaluation on MVTec LOCO, in which AnomalyCLIP, AdaCLIP and WinCLIP reach 61.6\%, 60.9\% and 53.2\% image-level AUROC, not reported separately for logical anomalies~\citep{fucka2026objectcore}. The order also changes: LogSAD moves ahead of CSAD and SALAD, so in our comparison clean AUROC does not predict the ranking under capture shifts.

In contrast, our framework is largely unaffected: AUROC decreases by 0.7 points for Ours (no overlap), from 93.2\% to 92.5\%, and by 0.8 points for Ours (full recipe), from 94.9\% to 94.1\%, corresponding to relative drops of 1.6\% and 1.8\%. The same holds in every category (Table~\ref{tab:per_category}): the heterogeneous AUROC of our framework stays within 2.4 points of its clean value, while CSAD and SALAD lose 12.6 to 41.3 points. As explained in Section~\ref{sec:augmentation}, the robustness claim rests on Ours (no overlap).

On clean images, however, Ours (full recipe) is 1.5 to 2.8 points below SALAD and CSAD, and below the three methods in the bottom block of Table~\ref{tab:main}, SALAD$^\dagger$~\citep{fucka2025salad}, PSAD~\citep{kim2024psad} and SAM-LAD~\citep{peng2025samlad}, whose robustness we could not measure (\ref{app:baselines}). The gap to SALAD and CSAD comes mainly from Screw Bag and Splicing Connectors (Table~\ref{tab:per_category}).

\subsection{Robustness Analysis}
\label{sec:robustness}

A performance drop shows how much a method degrades, but not why. Under reduced brightness, degradation can arise in two distinct ways, each requiring a different remedy. First, the method may no longer distinguish darkened anomalies from darkened normal images, even when evaluated only on darkened images (a within-condition drop). Second, it may preserve this ranking within darkened images, but shift all scores upward. In that case, a darkened normal image may receive a higher score than a clean anomaly, preventing a single threshold from separating the mixed test set despite good within-condition ranking (a cross-condition drop). Per-condition AUROC (Table~\ref{tab:decomposition}; Section~\ref{sec:evaluation}) distinguishes these effects: its difference from clean AUROC quantifies the within-condition drop, whereas its difference from heterogeneous AUROC quantifies the cross-condition drop.

\begin{table}[H]
\centering
\footnotesize\setlength{\tabcolsep}{4pt}
\caption{\textbf{Decomposition of the drop.} Image-level logical AUROC (\%), five-category mean. Per-condition: the mean AUROC of the six perturbed conditions, each scored on its own (Table~\ref{tab:per_condition} gives each condition). Within-condition drop = clean $-$ per-condition: separation lost even within one condition. Cross-condition drop = per-condition $-$ heterogeneous: lost mainly because the scores move between conditions. The two add up to the drop of Table~\ref{tab:main} (Section~\ref{sec:evaluation}); for each normal-reference method, the larger one is in bold. Worst condition: the perturbed condition with the lowest five-category mean (B40: brightness $-40$\%; Z10: zoom-out 10\%; Z15: zoom-out 15\%).}
\label{tab:decomposition}
\begin{tabular*}{\textwidth}{@{\extracolsep{\fill}}>{\raggedright\arraybackslash}p{2.8cm}cccccr@{\ }l@{}}
\toprule
& & Per- & Hetero- & Within-condition & Cross-condition & \multicolumn{2}{c}{Worst} \\
Method & Clean & condition & geneous & drop & drop & \multicolumn{2}{c}{condition} \\
\midrule
\textbf{Ours (no overlap)} & 93.2 & 92.8 & 92.5 & 0.4 & 0.3 & 91.0 & Z15 \\
\textbf{Ours (full recipe)} & 94.9 & 94.1 & 94.1 & 0.8 & 0.0 & 93.0 & Z15 \\
\midrule
CSAD & 97.7 & 75.0 & 68.1 & \textbf{22.7} & 6.9 & 61.9 & Z15 \\
SALAD & 96.4 & 85.6 & 62.3 & 10.8 & \textbf{23.3} & 78.0 & B40 \\
LogSAD & 89.2 & 77.5 & 73.9 & \textbf{11.7} & 3.6 & 67.6 & Z10 \\
PatchCore & 75.4 & 68.3 & 65.7 & \textbf{7.1} & 2.6 & 57.3 & Z15 \\
\midrule
AnomalyCLIP & 57.3 & 54.8 & 54.3 & 2.5 & 0.5 & 53.5 & Z15 \\
\bottomrule
\end{tabular*}
\end{table}

CSAD and SALAD fail in these two ways. Most of the drop of CSAD is within-condition drop (22.7 of 29.6 points): CSAD separates normal from anomalous images worse even within one condition. Most of the drop of SALAD is cross-condition drop (23.3 of 34.1 points): SALAD's scores move with the condition. For our framework, however, both parts stay below one point.

The difference matters: only the cross-condition drop could be repaired by per-condition recalibration. Such a recalibration keeps the order of the scores within each condition, so it cannot change the per-condition AUROC, which stays at 75.0\%, 85.6\% and 77.5\% for CSAD, SALAD and LogSAD, below our 92.8\% and 94.1\%. Our worst condition, the 15\% zoom-out (91.0\% and 93.0\%), is also above the worst condition of every baseline.

\begin{figure}[t]
    \centering
    \includegraphics[width=0.9\linewidth]{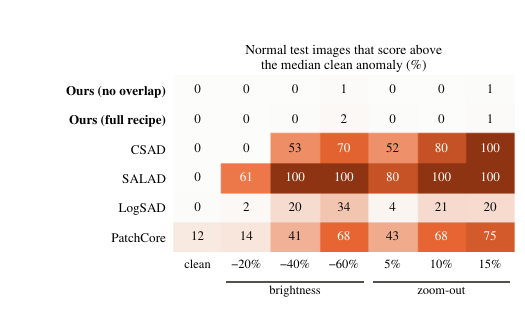}
\caption{\textbf{Normal images scoring above the median clean anomaly.} Each cell gives the share of normal (good) test images whose anomaly score is higher than the median score of the clean anomalous test images, as a five-category mean (\%). The clean column is the starting point: a detector that separates clean images well starts near 0\%, while PatchCore, which separates them less well, already starts at 12\%. A detector whose normal scores do not move with the capture conditions keeps its clean value in every column; a rise means that a brightness reduction or a zoom-out alone moves normal images above the median clean anomaly. AnomalyCLIP is left out: near chance on clean images, it already scores 38\% of its normal images above its median clean anomaly.}
\label{fig:score_shift}
\end{figure}

Fig.~\ref{fig:score_shift} shows how far the scores move with the condition: at the stronger settings, every normal-reference method we tested scores many normal test images above the median clean anomaly, while the classifier keeps almost all of them below it. Section~\ref{sec:why_robust} discusses possible causes of this difference.

\subsection{Ablation Study}
\label{sec:ablation}

\begin{table}[H]
\centering
\footnotesize\setlength{\tabcolsep}{4pt}
\caption{\textbf{Ablation.} Image-level logical AUROC (\%), mean over the five categories; each row below the first removes one component from the full recipe.}
\label{tab:ablation}
\begin{tabular}{lcc}
\toprule
Variant & Clean & Heterogeneous \\
\midrule
Ours (full recipe) & 94.9 & 94.1 \\
Without normal sub-mode discovery & 92.9 & 91.7 \\
Ours (no overlap): without test-overlapping augmentations & 93.2 & 92.5 \\
\bottomrule
\end{tabular}
\end{table}

Table~\ref{tab:ablation} removes one component at a time from the full recipe. Normal sub-mode discovery adds 2.0 points on clean images and 2.4 under the heterogeneous perturbation protocol, and changes only Juice Bottle and Splicing Connectors, the two categories that it splits (Section~\ref{sec:submodes}; per category in \ref{app:submodes}). Because this comparison keeps the full augmentation, we interpret it as a gain in accuracy, not as evidence of robustness. The test-overlapping augmentations add 1.7 points on clean images and 1.6 under the heterogeneous perturbation protocol; without them, as in Ours (no overlap), our framework is less accurate but still robust.
\section{Discussion}
\label{sec:discussion}

\subsection{Source of Robustness}
\label{sec:why_robust}

This section discusses possible reasons why the normal-reference methods we tested degrade substantially under the heterogeneous perturbation protocol, whereas our framework is largely unaffected (Section~\ref{sec:main_results}). Normal-reference detectors compare test images with normal data captured under a single lighting condition and camera position. We hypothesize that, as a result, reduced brightness or zooming out may increase the anomaly score of a defect-free product, making it difficult to distinguish a logical anomaly from a capture-related shift. Out-of-distribution detection makes a similar distinction between semantic shifts and covariate shifts, which change appearance without altering semantic content~\citep{yang2023fsood}. In these terms, a logical anomaly corresponds to a semantic shift and a capture shift to a covariate shift.

\noindent\textbf{Evidence.}
Fig.~\ref{fig:score_shift} supports this hypothesis: at the strongest settings, CSAD and SALAD score most normal test images above the median clean anomaly. A fully discriminative decision behaves differently: Ours (no overlap) retains its AUROC although neither perturbation occurs in its training, almost none of its normal test images score above the median clean anomaly (Fig.~\ref{fig:score_shift}), and both parts of its drop are close to zero (Section~\ref{sec:robustness}).

\noindent\textbf{Candidate explanations.}
These observations show that our decision is robust, but do not explain why. One possibility is that the score is relative: it weighs evidence for normal classes against evidence for anomaly types learned from images captured under the same conditions. In contrast, a distance-to-normal score may increase with any appearance change. Remaining training augmentations, such as blur and noise, may also contribute.

\noindent\textbf{Two failure modes.}
The two failure modes described in Section~\ref{sec:robustness} have different practical implications. When performance loss is primarily a cross-condition drop, as for SALAD, the method largely preserves its ranking within each condition. The main issue is that a single decision threshold no longer works across conditions. This can be addressed outside the detector by monitoring lighting and camera placement and applying condition-specific thresholds.

When the loss is primarily a within-condition drop, as for CSAD, condition-specific thresholds cannot restore performance (Section~\ref{sec:robustness}). In this case, the practical safeguard is to maintain capture conditions as closely as possible to those used in training.

\noindent\textbf{Test-time adaptation.}
Both safeguards operate outside a trained detector; test-time adaptation (Section~\ref{sec:related_robustness}) instead adapts to the test images and could be applied to our framework as well. Our results address a prior question: whether the decision itself is robust without adaptation.

\subsection{Limitations and Open Questions}
\label{sec:limitations}

\noindent\textbf{Clean AUROC.}
On clean images, the full recipe is 1.5 to 2.8 points below SALAD and CSAD (Section~\ref{sec:main_results}). On a line where capture is held as tightly as in training, a normal-reference method may therefore be the better choice.

\noindent\textbf{Human input.}
A person writes the short description of each category, which defines what counts as normal, and checks what is generated from it. Automating both steps is future work.

\noindent\textbf{Synthetic anomalies.}
The description also limits the anomaly types available for training: types derived from it cannot guarantee that every real anomaly is covered, and we did not evaluate performance separately on real anomalies of types outside the generated list. Generation artifacts, such as blending seams, could act as shortcuts; because the test set contains only real images, any harm from them is already reflected in the reported AUROC.

\noindent\textbf{Scope of the evaluation.}
Our evaluation is narrower than the setting that motivates it. Our test perturbations are synthetic and come from two families, brightness reduction and zoom-out (Section~\ref{sec:evaluation} gives the reasons). Real low-light capture also adds sensor noise and uneven illumination, and the object may shift relative to the camera; our perturbations simulate neither. Robustness to wider and real capture shifts remains open.

\noindent\textbf{Closed models.}
The language model and the image editor are closed models, so the generation stage cannot be repeated exactly; the material we will release is listed in the data availability statement.

\noindent\textbf{Role of the anomaly design.}
Our results evaluate the framework as a whole and do not isolate the contribution of designing anomalies with a language model. Whether this design outperforms simpler synthesis, such as cut-and-paste anomalies~\citep{li2021cutpaste}, remains open and would require a controlled comparison with the rest of the framework unchanged.
\section{Conclusion}
\label{sec:conclusion}

We showed that the normal-reference detectors are fragile under two simple capture shifts. Under the heterogeneous perturbation protocol, they lose 9.7 to 34.1 AUROC points, 38.2\% to 73.5\% of their performance above chance. They degrade in two ways: some lose discrimination within a condition, which no per-condition recalibration can repair, and others shift their scores between conditions.

We proposed a framework in which a classifier is trained on normal images and synthetic anomalies generated from brief category descriptions, then scores test images without reference to normal data. Anomaly types derived from a short description of each category are applied to normal images with an instruction-based image editor, and the classifier learns to separate the edited images from automatically discovered normal sub-modes. With its full training recipe, the framework reaches 94.9\% AUROC on clean images, 1.5 to 2.8 points below SALAD and CSAD, and 94.1\% under the perturbation protocol; trained without brightness or zoom-out augmentation, it loses only 0.7 points (93.2\% to 92.5\%).

These results suggest that clean AUROC alone may not predict how a detector behaves on a production line, and that a change in lighting or camera position alone can cause normal-reference detectors to score defect-free products as anomalous. Our evaluation, however, covers one benchmark and two synthetic shift families, and a person still writes the category description and checks what is generated from it. Future work includes automating these two steps, testing wider and real capture shifts, localizing the detected anomalies and extending the framework to structural anomalies.

\section*{Declaration of competing interests}
The authors declare that they have no known competing financial interests or personal relationships that could have appeared to influence the work reported in this paper.

\section*{Funding}
This work was supported by the Scientific and Technological Research Council of T\"{u}rkiye (T\"{U}B\.{I}TAK) under the B\.{I}DEB 2211 National PhD Scholarship Program.

\section*{CRediT authorship contribution statement}
\textbf{Ufuk Efe:} Conceptualization, Methodology, Software, Investigation, Data curation, Visualization, Writing (original draft).
\textbf{Alptekin Temizel:} Methodology, Supervision, Writing (review and editing).

\section*{Data availability}

The evaluation is conducted on the publicly available MVTec LOCO AD dataset~\citep{bergmann2022beyond}. The category descriptions, anomaly types and editing prompts, the curated synthetic training images and the code used in this study will be made publicly available upon acceptance of the manuscript; the synthetic images, which are derived from MVTec LOCO AD, will be released under its CC BY-NC-SA 4.0 license.

\section*{Declaration of generative AI and AI-assisted technologies in the manuscript preparation process}

During the preparation of this work the authors used Claude (Anthropic) to assist with drafting, editing and checking the manuscript text, and Claude (Anthropic) and Codex (OpenAI) to assist with writing and checking the analysis code. After using these tools, the authors reviewed and edited the content as needed and take full responsibility for the content of the published article. The language model and the image editor used to generate synthetic training data are part of the proposed method and are described in Section~\ref{sec:method}; the synthetic anomalies shown in Fig.~\ref{fig:pipeline}, Fig.~\ref{fig:examples} and Figs.~\ref{fig:examples_bb}--\ref{fig:examples_sc} were generated by this image editor.

\appendix
\counterwithin*{table}{section}
\counterwithin*{figure}{section}
\counterwithin*{equation}{section}
\sloppy

\newcommand{\apair}[3][0.23]{%
  \begin{minipage}{#1\textwidth}\centering
    \includegraphics[width=0.48\linewidth]{#2_good_#3.jpg}%
    \hfill
    \includegraphics[width=0.48\linewidth]{#2_synth_#3.jpg}%
    \\[-5pt]{\scriptsize \##3}%
    \vspace{4pt}
  \end{minipage}%
}
\newcommand{\anopair}[3][0.23]{%
  \begin{minipage}{#1\textwidth}\centering
    \sbox0{\includegraphics[width=0.48\linewidth]{#2_good_1.jpg}}%
    \begin{minipage}[c][\ht0][c]{\linewidth}\centering{\scriptsize \##3: not used}\end{minipage}%
    \\[-5pt]{\scriptsize \phantom{\##3}}%
    \vspace{4pt}
  \end{minipage}%
}
\newenvironment{promptbox}{\begin{quote}\singlespacing\small\sloppy}{\end{quote}}

\section{Dataset statistics}
\label{app:data}
\begingroup
\let\origtable\table \let\endorigtable\endtable
\renewenvironment{table}[1][]{\origtable[H]}{\endorigtable}
\begin{table}[t]
\centering
\footnotesize\setlength{\tabcolsep}{4pt}
\caption{\textbf{Data per category.} Real MVTec LOCO normal (good) images of the train, validation and test splits and the real logical test anomalies; synthetic anomalies used for training and validation; the number of normal sub-modes $K$; and the number of the 20 designed anomaly types used in training.}
\label{tab:data}
\begin{tabular*}{\textwidth}{@{\extracolsep{\fill}}lcccccccc@{}}
\toprule
& \multicolumn{3}{c}{Good} & Test & \multicolumn{2}{c}{Synthetic} & & \\
\cmidrule(lr){2-4}\cmidrule(lr){6-7}
Category & Train & Val & Test & logical & Train & Val & $K$ & Types \\
\midrule
Breakfast Box & 351 & 62 & 102 & 83 & 311 & 55 & 1 & 19 \\
Juice Bottle & 335 & 54 & 94 & 142 & 329 & 53 & 4 & 20 \\
Pushpins & 372 & 69 & 138 & 91 & 273 & 51 & 1 & 17 \\
Screw Bag & 360 & 60 & 122 & 137 & 258 & 45 & 1 & 17 \\
Splicing Connectors & 360 & 60 & 119 & 108 & 292 & 48 & 3 & 18 \\
\bottomrule
\end{tabular*}
\end{table}
\endgroup

Table~\ref{tab:data} lists the number of images per category and split. The synthetic columns count the synthetic anomalies kept after curation (Section~\ref{sec:generation}), and $K$ is the number of normal sub-modes (\ref{app:submodes}). An anomaly type becomes a class only if curation keeps at least eight of its training images and at least one of its validation images; ``Types'' counts these classes ($N_c$), and the other types are marked ``(not used)'' in Tables~\ref{tab:anomalies_bb}--\ref{tab:anomalies_sc}.

\section{Training details}
\label{app:training}

This appendix provides the details needed to reproduce the training of the classifier. Table~\ref{tab:hyper} lists the training settings, which are the same for all categories, and Table~\ref{tab:augment} the augmentation pipeline. Training runs with PyTorch's deterministic algorithms enabled, so a run is reproducible on the same hardware and software. One run takes about 40 minutes on one NVIDIA GeForce RTX 3080 (12\,GB), with PyTorch 2.9.1, timm 1.0.22~\citep{wightman2019timm} and Albumentations 2.0.8~\citep{buslaev2020albumentations}.

\begin{table}[H]
\centering
\footnotesize
\caption{\textbf{Training settings of the classifier (all categories).}}
\label{tab:hyper}
\begin{tabular}{@{}l>{\raggedright\arraybackslash}p{0.74\textwidth}@{}}
    \toprule
    Setting & Value \\
    \midrule
    Backbone & ConvNeXt-L~\citep{liu2022convnext}, DINOv3 LVD-1689M weights (\texttt{timm}: \texttt{convnext\_large.dinov3\_lvd1689m}; \citealp{simeoni2025dinov3}) \\
    Fine-tuning & full network, no frozen layers \\
    Head & global average pooling, LayerNorm, linear layer \\
    Outputs & $K$ normal sub-modes $+$ 20 anomaly types (21 outputs; Juice Bottle 24, Splicing Connectors 23); a type not used as a class keeps its output but has no training images \\
    Input & resized to $448 \times 448$, ImageNet mean and standard deviation \\
    Optimizer & AdamW~\citep{loshchilov2019adamw}, $\beta = (0.9, 0.999)$, $\epsilon = 10^{-8}$, weight decay 0.05 \\
    Learning rate & $10^{-5}$ for backbone and head, constant (no schedule) \\
    Batch & 2 images, gradient accumulation over 2 steps (effective batch 4) \\
    Epochs & 30 \\
    Regularization & no label smoothing, no dropout, no stochastic depth \\
    Checkpoint & lowest cross-entropy on the validation split (real normal images and synthetic anomalies) \\
    Seed & 42 \\
    \bottomrule
\end{tabular}
\end{table}

\begin{table}[H]
\centering
\footnotesize
\caption{\textbf{Augmentation pipeline (Albumentations), applied in this order to normal images and synthetic anomalies alike.} A ``one of'' block applies one of its members with the given probability. The blocks marked * contain the test perturbations and are removed in the variant without test-overlapping augmentations. Validation and test images are only resized and normalized.}
\label{tab:augment}
\begin{tabular}{@{}>{\raggedright}p{0.25\textwidth}c>{\raggedright\arraybackslash}p{0.57\textwidth}@{}}
    \toprule
    Block & $p$ & Operators and ranges \\
    \midrule
    Flips and rotation & 0.5 each & horizontal flip, vertical flip, rotation by a multiple of $90^{\circ}$ \\
    Zoom-out* & 0.3 & downscale by 0--15\%, edge-replicate padding back to $448 \times 448$ \\
    Blur (one of) & 0.5 & Gaussian ($\sigma$ 1--5), defocus (radius 2--3), motion (kernel 3--7), zoom blur (factor up to 1.05) \\
    Noise (one of) & 0.5 & per-pixel multiplicative noise (0.7--1.3), Gaussian noise (std 0.01--0.05), JPEG compression, downscaling by 0.5 \\
    Photometric* (one of) & 0.5 & brightness ($-60\%$ to $+40\%$ of the image mean) and contrast ($\pm 0.2$), CLAHE (clip limit 1--4), random tone curve, sharpening \\
    Color (one of) & 0.25 & emboss, posterize (4 bits) \\
    \bottomrule
\end{tabular}
\end{table}

\section{Prompts and generation settings}
\label{app:generation}

This appendix traces the synthetic anomalies from the category description to the edited image (Fig.~\ref{fig:pipeline}). A person writes a free-text description of the category, and the language model converts it into a structured description (Step~1a), proposes anomaly types based on that description (Step~1b) and writes one editing prompt per type (Step~1c). The image editor (\texttt{gemini-3-pro-image-preview}, accessed through Vertex AI) then applies the prompts (Step~2). The instructions differ slightly between categories; the differences are marked below.

\paragraph{Step 1a: category description}
For each category, a person wrote a short free-text description of the normal images. The language model converted this text into a structured description with the following fixed instruction. Only this structured description is passed on to Steps~1b and~1c; all such descriptions are given in~\ref{app:descriptions}.
\begin{promptbox}\itshape
Using the given dataset description, create a structured dataset description. It should be written so that it generalizes. It should be in English. Keep it plain and clear but complete. I only want the dataset description; do not add unnecessary detail. \textnormal{[Screw Bag and Splicing Connectors only:]} Below is an example of a good description written for another dataset. \textnormal{[the structured description of another category]}
\end{promptbox}

\paragraph{Logical anomaly definition}
The following definition, based on \citet{bergmann2022beyond}, is part of the Step~1b and Step~1c prompts:
\begin{promptbox}\itshape
Logical anomaly: Violations of the logical or geometrical constraints that govern component presence and relations, e.g., permissible objects in invalid locations, missing/extra parts, or incorrect arrangements. These depend on long-range dependencies and global composition rather than local textures, requiring reasoning over counts, positions, and relationships.
\end{promptbox}

\paragraph{Step 1b: anomaly types}
The structured description and the definition are passed with the following prompt (spelling corrected):
\begin{promptbox}\itshape
Considering the provided structured dataset description and the definition of a logical anomaly, generate a clear, concise, and specific list of 20 logical anomaly cases that could appear in this dataset. Each anomaly should be stated as a short, explicit sentence. \textnormal{[Not in the Juice Bottle version:]} Each anomaly should be clear, should not include ``or'' statements. Each anomaly should be independent from each other, i.e.\ one anomaly type cannot be produced by another anomaly type.

\medskip
\textnormal{[Splicing Connectors only:]} IMPORTANT (splicing\_connectors specific): Write anomalies in a type-agnostic way where possible. For example: Instead of ``T2 connector replaced with T3'' $\rightarrow$ ``One connector is a different type than the other''. Instead of ``Yellow cable used with T3 connectors'' $\rightarrow$ ``Cable color does not match the connector type''. Use relative terms: ``fewer ports'', ``more ports'', ``different color'', ``wrong type''. This is because the input image type (T2/T3/T5) is unknown at generation time.

\medskip
DATASET DESCRIPTION: \textnormal{[structured category description]}

\medskip
LOGICAL ANOMALY DEFINITION: \textnormal{[definition above]}
\end{promptbox}

\paragraph{Step 1c: editing prompts}
The 20 types are converted into editing instructions with the following prompt:
\begin{promptbox}\itshape
ROLE: You turn anomaly statements into short image edit prompts.

\medskip
YOU WILL RECEIVE: 1) DATASET DESCRIPTION 2) LOGICAL ANOMALY DEFINITION 3) POSSIBLE ANOMALIES LIST (numbered)

\medskip
TASK: Convert EACH anomaly item into ONE straightforward, minimal edit prompt for an AI image editing tool that modifies the given image to create exactly that anomaly. \textnormal{[Not in the Juice Bottle version:]} These prompts will be given to an AI image editor along with a normal (good) reference image from the dataset, with the goal of transforming that image into an anomalous one using the prompt.

\medskip
RULES: Keep the same numbering as the input list. One sentence per item. State only the change needed. Do not give specific examples.

\medskip
\textnormal{[Splicing Connectors only:]} ADDITIONAL RULES (splicing\_connectors specific): Since the input image's connector type (T2/T3/T5) is unknown at generation time, prompts must be type-agnostic. Use relative terms: ``a different type'', ``fewer ports'', ``more ports'', ``the wrong color''. Do NOT use specific values: ``T2'', ``T3'', ``T5'', ``yellow'', ``blue'', ``red''. Example: Instead of ``Change cable to yellow'' $\rightarrow$ ``Change the cable to a color that does not match the connector type''. Example: Instead of ``Replace with T5 connector'' $\rightarrow$ ``Replace one connector with a connector that has more ports''.

\medskip
DATASET DESCRIPTION: \textnormal{[structured category description]}

\medskip
LOGICAL ANOMALY DEFINITION: \textnormal{[definition above]}

\medskip
POSSIBLE ANOMALIES LIST: \textnormal{[the 20 types from Step 1b]}
\end{promptbox}

\paragraph{Step 2: image editing}
The image editor receives the editing prompt and the image and returns one candidate with its default sampling settings (temperature 1.0, top-$p$ 0.95, top-$k$ 64); the output is resized to the input resolution. This step is not exactly reproducible, because the image editor is a closed preview model, so we will release the category descriptions, prompts and synthetic anomalies.

\subsection{Category descriptions}
\label{app:descriptions}

These are the structured descriptions produced in Step~1a, as passed to the language model in Steps~1b and~1c (only the formatting is adapted). They define what a normal image is, and the anomaly types of~\ref{app:types} are derived from them as violations.

\paragraph{Breakfast Box}
\begin{promptbox}
\textbf{Dataset Content:} A collection of images showing a two-compartment breakfast tray containing food items arranged in a consistent layout.

\textbf{Tray Characteristics}
\begin{itemize}\itemsep0pt
  \item The tray is always a \textbf{white foam tray} with \textbf{two compartments}.
  \item The \textbf{left compartment} is consistently \textbf{narrower}, and the \textbf{right compartment} is consistently \textbf{wider}.
\end{itemize}
\textbf{Left Compartment Content (Fixed Set, Variable Order)}
\begin{itemize}\itemsep0pt
  \item The left compartment contains \textbf{exactly three whole fruits} in every image: \textbf{2 oranges}, \textbf{1 peach}.
  \item The fruits are placed in a \textbf{single vertical column} (top-to-bottom, one fruit per position).
  \item The \textbf{order is variable} across images (any permutation of two oranges and one peach is allowed).
  \item No other items appear in the left compartment.
\end{itemize}
\textbf{Right Compartment Content (Fixed Regions).} The right compartment is consistently divided into two content regions:
\begin{enumerate}\itemsep0pt
  \item \textbf{Upper Region (Cornflakes):} Contains \textbf{only cornflakes}. Cornflakes occupy \textbf{more than half} of the right compartment's area. The cornflakes form a \textbf{single contiguous, unmixed region} (no other items present within this upper area).
  \item \textbf{Lower Region (Mixed Banana and Almonds):} Contains a \textbf{mixture of sliced banana and almonds}. Banana slices and almonds are \textbf{randomly interspersed} (no fixed pattern or separation). No other items appear in the lower region.
\end{enumerate}
\end{promptbox}

\paragraph{Juice Bottle}
\begin{promptbox}
\textbf{Dataset Content:} The dataset contains images of transparent bottles filled with juice.

\textbf{Bottle Characteristics}
\begin{itemize}\itemsep0pt
  \item All bottles are transparent and filled with juice.
  \item Bottles are nearly full, leaving a small empty space at the top.
  \item Bottles have no caps and their openings are uncovered.
\end{itemize}
\textbf{Labels on Bottles.} Each bottle has exactly two labels:
\begin{enumerate}\itemsep0pt
  \item \textbf{First Label (Fruit Icon Label):} Positioned at the center of the bottle. Contains an icon representing the type of fruit juice inside. The color of the juice inside the bottle corresponds with the fruit icon on this label. The fruit icon is centered on this label.
  \item \textbf{Second Label (``100\% Juice''):} Located at the bottom part of the bottle. Always contains the text ``100\% Juice''.
\end{enumerate}
\end{promptbox}

\paragraph{Pushpins}
\begin{promptbox}
\textbf{Dataset Content:} A collection of images showing a \textbf{transparent plastic organizer box} containing \textbf{pushpins} arranged in a fully regular compartment layout.

\textbf{Container Characteristics}
\begin{itemize}\itemsep0pt
  \item The container is always the \textbf{same type of transparent plastic box}.
  \item The box is partitioned into a \textbf{fixed $3 \times 5$ grid} of compartments: \textbf{3 rows}, \textbf{5 columns}, \textbf{15 compartments total}.
  \item Each compartment is a distinct \textbf{separate cell} (i.e., a divider-based section).
\end{itemize}
\textbf{Compartment Content (Fixed Count, Variable Placement)}
\begin{itemize}\itemsep0pt
  \item Every image contains \textbf{exactly one pushpin in each compartment}.
  \item All pushpins are: \textbf{Yellow}; \textbf{The same size}.
  \item The \textbf{pose and placement} of each pushpin within its compartment are \textbf{variable}: Orientation is random. Position within the compartment is random.
  \item No compartment contains \textbf{zero} pushpins or \textbf{more than one} pushpin.
  \item No additional objects appear in the compartments or the box.
\end{itemize}
\end{promptbox}

\paragraph{Screw Bag}
\begin{promptbox}
\textbf{Dataset Content:} A collection of images showing a \textbf{transparent plastic bag} containing a \textbf{fixed set of metallic hardware components}.

\textbf{Container Characteristics}
\begin{itemize}\itemsep0pt
  \item The container is always a \textbf{transparent plastic bag}.
  \item All components appear \textbf{inside the bag}.
\end{itemize}
\textbf{Bag Content (Fixed Set, Fixed Counts).} Each image contains \textbf{exactly} the following items inside the bag:
\begin{itemize}\itemsep0pt
  \item \textbf{2 nuts} (consistent standard size across all samples)
  \item \textbf{2 washers} (consistent standard size across all samples)
  \item \textbf{2 screws}: \textbf{1 long screw} (consistent standard length across all samples); \textbf{1 short screw} (consistent standard length across all samples)
\end{itemize}
\textbf{Variability Constraints}
\begin{itemize}\itemsep0pt
  \item The \textbf{positions} and \textbf{orientations} of all items \textbf{may vary arbitrarily} across images.
  \item The \textbf{item counts} and the \textbf{long vs.\ short screw length distinction} are \textbf{invariant} for all samples.
\end{itemize}
\end{promptbox}

\paragraph{Splicing Connectors}
\begin{promptbox}
\textbf{Dataset Content:} A collection of images showing \textbf{electrical cable splicing connectors} placed on a \textbf{metal grid background}.

\textbf{Scene Structure (Fixed Counts).} Each image contains \textbf{exactly} (no more, no less):
\begin{itemize}\itemsep0pt
  \item \textbf{2 splicing connectors} (no more, no less; one on the \textbf{left}, one on the \textbf{right})
  \item \textbf{1 cable} connecting the two connectors (no more, no less)
\end{itemize}
\textbf{Connector Types and Cable Color Mapping.} Connectors can be one of the following types (defined by the \textbf{number of ports/levers}):
\begin{itemize}\itemsep0pt
  \item \textbf{T2:} 2-port connector (small, \textbf{2 levers}) $\rightarrow$ \textbf{Yellow cable}
  \item \textbf{T3:} 3-port connector (medium, \textbf{3 levers}) $\rightarrow$ \textbf{Blue cable}
  \item \textbf{T5:} 5-port connector (large, \textbf{5 levers}) $\rightarrow$ \textbf{Red cable}
\end{itemize}
\textbf{Consistency Rules (Invariant Across All Normal Samples)}
\begin{itemize}\itemsep0pt
  \item The \textbf{two connectors in the same image must be the same type} (both T2, both T3, or both T5).
  \item The \textbf{cable color is strictly determined by connector type} (T2=yellow, T3=blue, T5=red), with \textbf{no exceptions}. \textbf{Only these 3 colors exist} in normal samples; cables of other colors never appear.
  \item The cable must be \textbf{inserted into both connectors} (i.e., physically connected on both ends).
  \item The cable must be connected to the \textbf{same lever index/row on both connectors} (the specific lever used may vary between images, but must match left-to-right).
  \item All connector levers must be \textbf{closed}.
  \item The connectors must be \textbf{upright and aligned}, facing each other in a straight configuration.
  \item The cable length must remain within a \textbf{standard expected range} (not excessively short or long).
  \item The cable must be \textbf{intact and continuous} (not broken or cut).
\end{itemize}
\textbf{Allowed Variations}
\begin{itemize}\itemsep0pt
  \item The \textbf{positions}, \textbf{spacing}, \textbf{orientation}, and \textbf{viewpoint} of the connectors and cable may vary between images, as long as all rules above remain satisfied.
\end{itemize}
\end{promptbox}

\subsection{Anomaly types, editing prompts and examples}
\label{app:types}

\begin{figure}[H]
\centering
\apair[0.31]{breakfast_box}{1}\hfill\apair[0.31]{breakfast_box}{2}\hfill\apair[0.31]{breakfast_box}{3}\\[1pt]
\apair[0.31]{breakfast_box}{4}\hfill\apair[0.31]{breakfast_box}{5}\hfill\apair[0.31]{breakfast_box}{6}\\[1pt]
\apair[0.31]{breakfast_box}{7}\hfill\apair[0.31]{breakfast_box}{8}\hfill\apair[0.31]{breakfast_box}{9}\\[1pt]
\apair[0.31]{breakfast_box}{10}\hfill\apair[0.31]{breakfast_box}{11}\hfill\anopair[0.31]{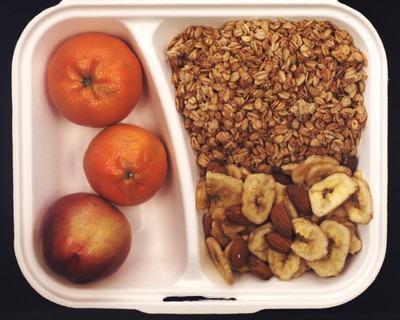}{12}\\[1pt]
\apair[0.31]{breakfast_box}{13}\hfill\apair[0.31]{breakfast_box}{14}\hfill\apair[0.31]{breakfast_box}{15}\\[1pt]
\apair[0.31]{breakfast_box}{16}\hfill\apair[0.31]{breakfast_box}{17}\hfill\apair[0.31]{breakfast_box}{18}\\[1pt]
\apair[0.31]{breakfast_box}{19}\hfill\apair[0.31]{breakfast_box}{20}\hfill\hspace*{0.31\textwidth}
\scriptsize
\caption{\textbf{Breakfast Box: example pairs.} Normal training images (left) and synthetic anomalies made from them (right). Numbers are the type IDs of Table~\ref{tab:anomalies_bb}. The synthetic anomalies are generated by the image editor (Gemini 3 Pro Image), a generative AI model (Section~\ref{sec:generation}).}
\label{fig:examples_bb}
\end{figure}

\begin{figure}[H]
\centering
\apair[0.31]{pushpins}{1}\hfill\apair[0.31]{pushpins}{2}\hfill\apair[0.31]{pushpins}{3}\\[1pt]
\apair[0.31]{pushpins}{4}\hfill\apair[0.31]{pushpins}{5}\hfill\apair[0.31]{pushpins}{6}\\[1pt]
\apair[0.31]{pushpins}{7}\hfill\anopair[0.31]{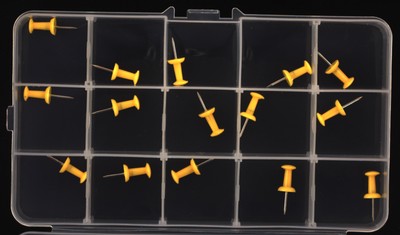}{8}\hfill\anopair[0.31]{pushpins}{9}\\[1pt]
\apair[0.31]{pushpins}{10}\hfill\apair[0.31]{pushpins}{11}\hfill\apair[0.31]{pushpins}{12}\\[1pt]
\apair[0.31]{pushpins}{13}\hfill\apair[0.31]{pushpins}{14}\hfill\apair[0.31]{pushpins}{15}\\[1pt]
\apair[0.31]{pushpins}{16}\hfill\anopair[0.31]{pushpins}{17}\hfill\apair[0.31]{pushpins}{18}\\[1pt]
\apair[0.31]{pushpins}{19}\hfill\apair[0.31]{pushpins}{20}\hfill\hspace*{0.31\textwidth}
\scriptsize
\caption{\textbf{Pushpins: example pairs.} Normal training images (left) and synthetic anomalies made from them (right). Numbers are the type IDs of Table~\ref{tab:anomalies_pp}. The synthetic anomalies are generated by the image editor (Gemini 3 Pro Image), a generative AI model (Section~\ref{sec:generation}).}
\label{fig:examples_pp}
\end{figure}

\begin{figure}[H]
\centering
\apair[0.31]{screw_bag}{1}\hfill\apair[0.31]{screw_bag}{2}\hfill\apair[0.31]{screw_bag}{3}\\[1pt]
\apair[0.31]{screw_bag}{4}\hfill\apair[0.31]{screw_bag}{5}\hfill\apair[0.31]{screw_bag}{6}\\[1pt]
\apair[0.31]{screw_bag}{7}\hfill\apair[0.31]{screw_bag}{8}\hfill\apair[0.31]{screw_bag}{9}\\[1pt]
\apair[0.31]{screw_bag}{10}\hfill\anopair[0.31]{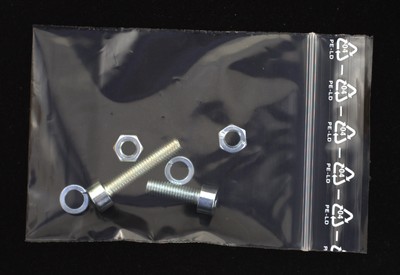}{11}\hfill\apair[0.31]{screw_bag}{12}\\[1pt]
\apair[0.31]{screw_bag}{13}\hfill\apair[0.31]{screw_bag}{14}\hfill\apair[0.31]{screw_bag}{15}\\[1pt]
\apair[0.31]{screw_bag}{16}\hfill\apair[0.31]{screw_bag}{17}\hfill\apair[0.31]{screw_bag}{18}\\[1pt]
\anopair[0.31]{screw_bag}{19}\hfill\anopair[0.31]{screw_bag}{20}\hfill\hspace*{0.31\textwidth}
\scriptsize
\caption{\textbf{Screw Bag: example pairs.} Normal training images (left) and synthetic anomalies made from them (right). Numbers are the type IDs of Table~\ref{tab:anomalies_sb}. The synthetic anomalies are generated by the image editor (Gemini 3 Pro Image), a generative AI model (Section~\ref{sec:generation}).}
\label{fig:examples_sb}
\end{figure}

\begin{figure}[H]
\centering
\apair[0.31]{splicing_connectors}{1}\hfill\apair[0.31]{splicing_connectors}{2}\hfill\apair[0.31]{splicing_connectors}{3}\\[1pt]
\apair[0.31]{splicing_connectors}{4}\hfill\apair[0.31]{splicing_connectors}{5}\hfill\apair[0.31]{splicing_connectors}{6}\\[1pt]
\apair[0.31]{splicing_connectors}{7}\hfill\anopair[0.31]{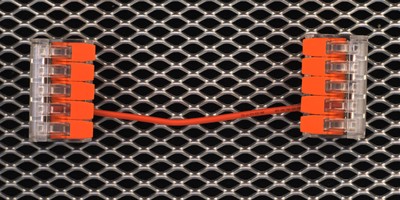}{8}\hfill\apair[0.31]{splicing_connectors}{9}\\[1pt]
\apair[0.31]{splicing_connectors}{10}\hfill\apair[0.31]{splicing_connectors}{11}\hfill\apair[0.31]{splicing_connectors}{12}\\[1pt]
\anopair[0.31]{splicing_connectors}{13}\hfill\apair[0.31]{splicing_connectors}{14}\hfill\apair[0.31]{splicing_connectors}{15}\\[1pt]
\apair[0.31]{splicing_connectors}{16}\hfill\apair[0.31]{splicing_connectors}{17}\hfill\apair[0.31]{splicing_connectors}{18}\\[1pt]
\apair[0.31]{splicing_connectors}{19}\hfill\apair[0.31]{splicing_connectors}{20}\hfill\hspace*{0.31\textwidth}
\scriptsize
\caption{\textbf{Splicing Connectors: example pairs.} Normal training images (left) and synthetic anomalies made from them (right). Numbers are the type IDs of Table~\ref{tab:anomalies_sc}. The synthetic anomalies are generated by the image editor (Gemini 3 Pro Image), a generative AI model (Section~\ref{sec:generation}).}
\label{fig:examples_sc}
\end{figure}

Tables~\ref{tab:anomalies_bb}--\ref{tab:anomalies_sc} list the 20 anomaly types of each category from Step~1b and their editing prompts from Step~1c, exactly as used for generation. Types marked ``(not used)'' fail the class rule of~\ref{app:data}. Figs.~\ref{fig:examples_bb}--\ref{fig:examples_sc} show, for every used type, a normal training image and a synthetic anomaly generated from that image with the type's prompt.

\begin{table}[H]
\centering
\scriptsize\setlength{\tabcolsep}{3pt}
\caption{\textbf{Breakfast Box: anomaly types and editing prompts.}}
\label{tab:anomalies_bb}
\begin{tabular}{@{}r p{0.41\textwidth} p{0.51\textwidth}@{}}
    \toprule
    ID & Anomaly type & Editing prompt \\
    \midrule
     1 & The tray is black. & Change the tray color to black. \\
     2 & The tray material is metal. & Change the tray material to metal. \\
     3 & The tray has three compartments. & Add a third compartment to the tray. \\
     4 & The left compartment is wider than the right compartment. & Resize the compartments so the left compartment is wider than the right compartment. \\
     5 & The left compartment contains two fruits. & Remove one fruit from the left compartment so it contains only two whole fruits. \\
     6 & The left compartment contains four fruits. & Add one extra whole fruit to the left compartment so it contains four whole fruits. \\
     7 & The left compartment contains one orange and two peaches. & Replace one orange in the left compartment with a peach so it contains one orange and two peaches. \\
     8 & The left compartment contains three oranges. & Replace the peach in the left compartment with an orange so it contains three oranges. \\
     9 & The left compartment contains an apple. & Add an apple to the left compartment. \\
    10 & The left compartment contains a peeled orange segment. & Replace one whole fruit in the left compartment with a peeled orange segment. \\
    11 & The left compartment contains a peeled peach segment. & Replace one whole fruit in the left compartment with a peeled peach segment. \\
    12 & Two fruits in the left compartment overlap within the same vertical position. \emph{(not used)} & Move one fruit in the left compartment to overlap another fruit at the same vertical position. \\
    13 & A whole orange is placed in the right compartment. & Add a whole orange into the right compartment. \\
    14 & Cornflakes appear in the lower region of the right compartment. & Add cornflakes into the lower region of the right compartment. \\
    15 & Banana slices appear in the upper cornflakes region. & Add banana slices into the upper cornflakes region of the right compartment. \\
    16 & Almonds are embedded inside the upper cornflakes region. & Add almonds embedded within the upper cornflakes region of the right compartment. \\
    17 & The cornflakes region occupies less than half of the right compartment area. & Reduce the cornflakes region so it covers less than half of the right compartment area. \\
    18 & The cornflakes region is split into two disconnected patches. & Split the cornflakes region into two disconnected patches. \\
    19 & The lower region contains banana slices without almonds. & Remove all almonds from the lower region so it contains only banana slices. \\
    20 & The lower region contains almonds without banana slices. & Remove all banana slices from the lower region so it contains only almonds. \\
    \bottomrule
\end{tabular}
\end{table}

\begin{table}[H]
\centering
\scriptsize\setlength{\tabcolsep}{3pt}
\caption{\textbf{Juice Bottle: anomaly types and editing prompts.}}
\label{tab:anomalies_jb}
\begin{tabular}{@{}r p{0.41\textwidth} p{0.51\textwidth}@{}}
    \toprule
    ID & Anomaly type & Editing prompt \\
    \midrule
     1 & Bottle is opaque. & Make the bottle opaque. \\
     2 & Bottle contains no juice (empty or water). & Remove the juice, leaving the bottle empty. \\
     3 & Liquid level is far below `nearly full'. & Lower the liquid level to well below nearly full. \\
     4 & Bottle is completely full with no headspace. & Fill the bottle to the brim, removing all headspace. \\
     5 & Headspace appears at the bottom or side. & Relocate the headspace to the bottom of the bottle. \\
     6 & Bottle has a cap attached. & Add a cap to the bottle. \\
     7 & Bottle opening is covered by a seal/film/cork. & Cover the bottle opening with a seal. \\
     8 & Bottle has more than two labels. & Add an extra label to the bottle. \\
     9 & Bottle has only one label. & Remove one label so only a single label remains. \\
    10 & Fruit icon label is not at the center of the bottle. & Move the fruit icon label away from the center of the bottle. \\
    11 & Fruit icon label lacks any fruit icon. & Remove the fruit icon from the fruit icon label. \\
    12 & Fruit icon is off-center within its label. & Shift the fruit icon within its label off-center. \\
    13 & Fruit icon depicts a different fruit than the juice. & Change the fruit icon to depict a different fruit than the juice. \\
    14 & Juice color does not match the fruit icon. & Change the juice color so it does not match the fruit icon. \\
    15 & `100\% Juice' label is not at the bottom. & Move the ``100\% Juice'' label away from the bottom. \\
    16 & Bottom label text differs from exactly `100\% Juice'. & Replace the bottom label text with wording other than ``100\% Juice.'' \\
    17 & `100\% Juice' label is missing. & Remove the ``100\% Juice'' label. \\
    18\textsuperscript{a} & Both labels say `100\% Juice' (no fruit icon label present). & Remove two labels so no labels remain. \\
    19 & Label roles are swapped (fruit icon at bottom, `100\% Juice' at center). & Swap the positions of the fruit icon label and the ``100\% Juice'' label. \\
    20 & Fruit icon label shows a non-fruit graphic (e.g., brand logo). & Replace the fruit icon with a non-fruit graphic on its label. \\
    \bottomrule
    \multicolumn{3}{@{}p{0.95\textwidth}@{}}{\textsuperscript{a}\,Type 18: the prompt removes both labels instead of following the statement; the class follows the prompt.} \\
\end{tabular}
\end{table}

\begin{figure}[H]
\centering
\apair[0.235]{juice_bottle}{1}\hfill\apair[0.235]{juice_bottle}{2}\hfill\apair[0.235]{juice_bottle}{3}\hfill\apair[0.235]{juice_bottle}{4}\\[1pt]
\apair[0.235]{juice_bottle}{5}\hfill\apair[0.235]{juice_bottle}{6}\hfill\apair[0.235]{juice_bottle}{7}\hfill\apair[0.235]{juice_bottle}{8}\\[1pt]
\apair[0.235]{juice_bottle}{9}\hfill\apair[0.235]{juice_bottle}{10}\hfill\apair[0.235]{juice_bottle}{11}\hfill\apair[0.235]{juice_bottle}{12}\\[1pt]
\apair[0.235]{juice_bottle}{13}\hfill\apair[0.235]{juice_bottle}{14}\hfill\apair[0.235]{juice_bottle}{15}\hfill\apair[0.235]{juice_bottle}{16}\\[1pt]
\apair[0.235]{juice_bottle}{17}\hfill\apair[0.235]{juice_bottle}{18}\hfill\apair[0.235]{juice_bottle}{19}\hfill\apair[0.235]{juice_bottle}{20}
\scriptsize
\caption{\textbf{Juice Bottle: example pairs.} Normal training images (left) and synthetic anomalies made from them (right). Numbers are the type IDs of Table~\ref{tab:anomalies_jb}. The synthetic anomalies are generated by the image editor (Gemini 3 Pro Image), a generative AI model (Section~\ref{sec:generation}).}
\label{fig:examples_jb}
\end{figure}

\begin{table}[H]
\centering
\scriptsize\setlength{\tabcolsep}{3pt}
\caption{\textbf{Pushpins: anomaly types and editing prompts.}}
\label{tab:anomalies_pp}
\begin{tabular}{@{}r p{0.41\textwidth} p{0.51\textwidth}@{}}
    \toprule
    ID & Anomaly type & Editing prompt \\
    \midrule
     1 & All compartments are empty; no pushpins are present in the box. & Remove all pushpins from every compartment so the entire box is empty. \\
     2 & The divider wall between two adjacent compartments is missing, merging them into one larger compartment. & Remove the divider wall between two adjacent compartments so they merge into one larger compartment. \\
     3 & The box has 15 compartments but they are not aligned into straight rows and columns. & Misalign the compartments so they are not in straight rows and columns. \\
     4 & One compartment contains no pushpin. & Remove the pushpin from one compartment leaving it empty. \\
     5 & Two compartments contain no pushpin. & Remove the pushpins from two compartments leaving them empty. \\
     6 & One compartment contains two pushpins. & Add an extra pushpin to one compartment so it contains two pushpins. \\
     7 & Two compartments contain two pushpins. & Add an extra pushpin to two compartments so each contains two pushpins. \\
     8 & One pushpin is located outside all compartments but still inside the box. \emph{(not used)} & Move one pushpin outside the box. \\
     9 & One pushpin is resting on top of a divider wall instead of inside a compartment. \emph{(not used)} & Place one pushpin on top of a divider wall instead of inside a compartment. \\
    10 & One compartment contains a paperclip instead of a pushpin. & Replace the pushpin in one compartment with a paperclip. \\
    11 & One compartment contains a pushpin together with a paperclip. & Add a paperclip next to the pushpin in one compartment. \\
    12 & One pushpin is red instead of yellow. & Change the color of one pushpin from yellow to red. \\
    13 & One pushpin is blue instead of yellow. & Change the color of one pushpin from yellow to blue. \\
    14 & One pushpin is green instead of yellow. & Change the color of one pushpin from yellow to green. \\
    15 & One compartment contains three pushpins instead of one. & Add two extra pushpins to one compartment so it contains three pushpins. \\
    16 & One pushpin is noticeably larger than the others. & Make one pushpin noticeably larger than the others. \\
    17 & One pushpin is noticeably smaller than the others. \emph{(not used)} & Make one pushpin noticeably smaller than the others. \\
    18 & One pushpin is missing its metal pin and only the head is present. & Remove the metal pin from one pushpin leaving only the head. \\
    19 & One pushpin is visibly bent into a curved shape. & Bend one pushpin into a curved shape. \\
    20 & Two pushpins are physically connected by a thread spanning across compartments. & Connect two pushpins in adjacent compartments with a thread. \\
    \bottomrule
\end{tabular}
\end{table}

\begin{table}[H]
\centering
\scriptsize\setlength{\tabcolsep}{3pt}
\caption{\textbf{Screw Bag: anomaly types and editing prompts.}}
\label{tab:anomalies_sb}
\begin{tabular}{@{}r p{0.41\textwidth} p{0.51\textwidth}@{}}
    \toprule
    ID & Anomaly type & Editing prompt \\
    \midrule
     1 & The long screw is significantly longer than its standard length. & Make the long screw significantly longer than its standard length. \\
     2 & The long screw is significantly shorter than its standard length. & Make the long screw significantly shorter than its standard length. \\
     3 & The short screw is significantly shorter than its standard length. & Make the short screw significantly shorter than its standard length. \\
     4 & The plastic bag is torn open while all six required components remain inside it. & Tear the plastic bag open while keeping all six components inside the bag. \\
     5 & Both screws are significantly shorter than their standard lengths. & Make both screws significantly shorter than their standard lengths. \\
     6 & Exactly one nut is inside the bag. & Remove one nut so that exactly one nut remains inside the bag. \\
     7 & Exactly three nuts are inside the bag. & Add one additional nut so that there are exactly three nuts inside the bag. \\
     8 & No nuts are inside the bag. & Remove both nuts so that no nuts are inside the bag. \\
     9 & Exactly four nuts are inside the bag. & Add two additional nuts so that there are exactly four nuts inside the bag. \\
    10 & Exactly one washer is inside the bag. & Remove one washer so that exactly one washer remains inside the bag. \\
    11 & Exactly three washers are inside the bag. \emph{(not used)} & Add one additional washer so that there are exactly three washers inside the bag. \\
    12 & No washers are inside the bag. & Remove both washers so that no washers are inside the bag. \\
    13 & Exactly four washers are inside the bag. & Add two additional washers so that there are exactly four washers inside the bag. \\
    14 & Only the long screw is inside the bag. & Remove the short screw, both nuts, and both washers so only the long screw remains inside the bag. \\
    15 & Only the short screw is inside the bag. & Remove the long screw, both nuts, and both washers so only the short screw remains inside the bag. \\
    16 & No screws are inside the bag. & Remove both screws so that no screws are inside the bag. \\
    17 & Three screws are inside the bag, including two long screws and one short screw. & Add one additional long screw so there are three screws inside the bag: two long and one short. \\
    18 & Three screws are inside the bag, including one long screw and two short screws. & Add one additional short screw so there are three screws inside the bag: one long and two short. \\
    19 & One nut is significantly smaller than its standard size. \emph{(not used)} & Make one nut significantly smaller than its standard size. \\
    20 & One washer is significantly smaller than its standard size. \emph{(not used)} & Make one washer significantly smaller than its standard size. \\
    \bottomrule
\end{tabular}
\end{table}

\begin{table}[H]
\centering
\scriptsize\setlength{\tabcolsep}{3pt}
\caption{\textbf{Splicing Connectors: anomaly types and editing prompts.}}
\label{tab:anomalies_sc}
\begin{tabular}{@{}r p{0.41\textwidth} p{0.51\textwidth}@{}}
    \toprule
    ID & Anomaly type & Editing prompt \\
    \midrule
     1 & Only one splicing connector is present. & Remove the right splicing connector so only one connector remains. \\
     2 & Two cables are present between the connectors. & Add a second cable connecting the two connectors. \\
     3 & No cable is present between the connectors. & Remove the cable so no cable is present between the connectors. \\
     4 & The cable is a color that is not used in normal samples. & Change the cable to a color that is not used in normal samples. \\
     5 & The two connectors are different types (different number of ports/levers). & Replace one connector with a connector that has a different number of ports/levers. \\
     6 & The cable color does not match the connector type. & Change the cable to a color that does not match the connector type. \\
     7 & The cable is inserted into the left connector but not inserted into the right connector. & Remove the cable from the right connector so it is only inserted into the left connector. \\
     8 & The cable is not inserted into either connector. \emph{(not used)} & Remove the cable from both connectors so it is not inserted into either connector. \\
     9 & The cable is inserted into different lever indices on the two connectors. & Move one end of the cable to a different lever index so the two ends are inserted into different lever indices. \\
    10 & At least one lever on a connector is open. & Open at least one lever on a connector. \\
    11 & A lever is missing or visibly broken on a connector. & Remove or visibly break a lever on a connector. \\
    12 & One connector is rotated so it is not upright. & Rotate one connector so it is not upright. \\
    13 & One connector is rotated so its ports face away from the other connector. \emph{(not used)} & Rotate one connector so its ports face away from the other connector. \\
    14 & The two connectors are not aligned to face each other in a straight configuration. & Reposition the connectors so they do not face each other in a straight configuration. \\
    15 & The cable is inserted into the same connector at both ends, leaving the other connector unconnected. & Insert both ends of the cable into the same connector and leave the other connector unconnected. \\
    16 & The two connectors are touching or overlapping. & Move the connectors so they touch or overlap. \\
    17 & The cable is excessively short relative to the standard expected range. & Shorten the cable so it is excessively short relative to the expected range. \\
    18 & The cable is excessively long relative to the standard expected range. & Lengthen the cable so it is excessively long relative to the expected range. \\
    19 & The cable is visibly cut, creating a break along its length. & Add a visible cut in the cable so it is broken along its length. \\
    20 & The cable connects to a connector position that is not a valid port/lever entry (i.e., attached to the connector body rather than a lever/port). & Attach one end of the cable to the connector body instead of a valid port/lever entry. \\
    \bottomrule
\end{tabular}
\end{table}

\section{Baseline setups}
\label{app:baselines}

This appendix lists the settings of each baseline, so that the robustness comparison can be repeated. All baselines use the official code and released weights without training any network (Section~\ref{sec:experiments}).

\paragraph{CSAD and SALAD}
Official code of CSAD (\url{https://github.com/Tokichan/CSAD}; \citealp{hsieh2024csad}) and SALAD (\url{https://github.com/MaticFuc/SALAD}; \citealp{fucka2025salad}) with the weights released by the authors, in their default settings.

\paragraph{LogSAD}
Official code of LogSAD (\url{https://github.com/zhang0jhon/LogSAD}; \citealp{zhang2025logsad}) in its full-data setting, with its frozen foundation models at $448 \times 448$. We score each image by the larger of LogSAD's two scores before its final sigmoid, keeping rule-flagged images on top, because the float64 sigmoid saturates at exactly 1.0 for large values and creates ties; where it does not saturate, the ranking and the AUROC are unchanged.

\paragraph{PatchCore}
Official code of PatchCore (\url{https://github.com/amazon-science/patchcore-inspection}; \citealp{roth2022patchcore}) with an ImageNet WideResNet-50-2, \texttt{layer2} and \texttt{layer3} features aggregated over $3 \times 3$ neighborhoods and pooled to 1024 dimensions, a 10\% greedy coreset of the clean training patches, and the largest nearest-neighbor patch distance as the image score. We resize the whole image to $256 \times 256$ instead of using the default center crop, which would cut off the ends of the non-square LOCO images.

\paragraph{AnomalyCLIP}
Official code of AnomalyCLIP (\url{https://github.com/zqhang/AnomalyCLIP}; \citealp{zhou2024anomalyclip}), run zero-shot with OpenAI CLIP ViT-L/14@336 and the prompt checkpoint the authors trained on VisA, at $518 \times 518$. The image score is the probability of the anomaly prompt; no LOCO image is used as a normal reference.

\paragraph{Methods with only a reported clean result}
SALAD$^\dagger$~\citep{fucka2025salad}, the setting of SALAD trained with the supervised part maps of PSAD, would need training; PSAD~\citep{kim2024psad} has no released MVTec LOCO weights and needs training with hand-labeled part masks; SAM-LAD~\citep{peng2025samlad} has no official implementation. We therefore report their published clean AUROC and could not measure their robustness.

\section{Discovering sub-modes of normal images}
\label{app:submodes}
\begin{table}[H]
\centering
\footnotesize\setlength{\tabcolsep}{4pt}
\caption{\textbf{Normal sub-mode discovery per category.} Only normal training images are used. The candidate $K$ is chosen by 5-fold cross-validated log-likelihood with the one-standard-error rule; a candidate must pass three gates (silhouette $\geq 0.25$, smallest cluster $\geq 10\%$, median bootstrap ARI $\geq 0.8$), otherwise $K-1$ is tried.}
\label{tab:submodes}
\begin{tabular*}{\textwidth}{@{\extracolsep{\fill}}lcp{0.52\textwidth}c@{}}
\toprule
Category & Candidate $K$ & Gates: silhouette / smallest cluster / ARI & Chosen $K$ \\
\midrule
Juice Bottle & 4 & $K{=}4$: 0.53 / 11\% / 0.94 pass & 4 \\
Splicing Connectors & 4 & $K{=}4$: 0.20 / 14\% / 0.94 fail (silhouette); $K{=}3$: 0.28 / 33\% / 1.00 pass & 3 \\
Screw Bag & 2 & $K{=}2$: 0.31 / 11\% / 0.66 fail (ARI) & 1 \\
Breakfast Box & 1 & single candidate, no gate & 1 \\
Pushpins & 1 & single candidate, no gate & 1 \\
\bottomrule
\end{tabular*}
\end{table}

This appendix gives the exact rule of normal sub-mode discovery (Section~\ref{sec:submodes}). It runs once per category before training and uses only the normal training images. The features are the pooled output of the frozen, pretrained ConvNeXt-L DINOv3 backbone at $448 \times 448$, normalized to unit length and reduced to eight dimensions by principal component analysis (PCA). Gaussian mixtures with a tied covariance are fitted for $K = 1, \dots, 5$, and the candidate is the smallest $K$ within one standard error of the best 5-fold cross-validated log-likelihood~\citep{hastie2009elements}. A candidate $K > 1$ must pass the three gates of Table~\ref{tab:submodes}, where the silhouette~\citep{rousseeuw1987silhouettes} is computed on the fitted clusters and the adjusted Rand index (ARI; \citealp{hubert1985comparing}) against 20 refits on bootstrap samples; otherwise $K-1, K-2, \dots$ are tried, down to $K = 1$. Finally, validation normal images are assigned to sub-modes by the mixture fitted on the training images.

Under this rule, in Juice Bottle, the banana juice images form two of the four clusters (75 and 37 images); in Splicing Connectors, each of the three clusters holds 120 images. Over 45 combinations of the gate thresholds (silhouette 0.15--0.35, smallest cluster 5--15\%, ARI 0.7--0.9), the same two categories are split; only the number of sub-modes varies.

With a single normal class instead of these sub-modes, the clean AUROC falls from 97.0\% to 92.2\% in Juice Bottle and from 90.3\% to 84.9\% in Splicing Connectors; the other categories are unchanged by construction.

\section{Results per condition}
\label{app:conditions}
\begin{table}[H]
\centering
\footnotesize\setlength{\tabcolsep}{4pt}
\caption{\textbf{Results per condition.} Image-level logical AUROC (\%), each condition scored alone, mean over the five categories. B20/B40/B60: brightness reduced by 20/40/60\%; Z5/Z10/Z15: zoom-out by 5/10/15\% with edge replication.}
\label{tab:per_condition}
\begin{tabular*}{\textwidth}{@{\extracolsep{\fill}}lccccccc@{}}
\toprule
Method & Clean & B20 & B40 & B60 & Z5 & Z10 & Z15 \\
\midrule
\textbf{Ours (no overlap)} & 93.2 & 93.2 & 93.4 & 93.8 & 92.8 & 92.4 & 91.0 \\
\textbf{Ours (full recipe)} & 94.9 & 94.8 & 94.7 & 94.4 & 94.1 & 93.6 & 93.0 \\
\midrule
CSAD & 97.7 & 92.2 & 82.0 & 74.4 & 74.7 & 64.8 & 61.9 \\
SALAD & 96.4 & 89.4 & 78.0 & 80.8 & 94.7 & 87.3 & 83.4 \\
LogSAD & 89.2 & 87.5 & 84.0 & 80.0 & 75.7 & 67.6 & 70.5 \\
PatchCore & 75.4 & 75.4 & 74.3 & 72.7 & 67.2 & 62.8 & 57.3 \\
\midrule
AnomalyCLIP & 57.3 & 55.3 & 55.1 & 55.4 & 54.3 & 55.5 & 53.5 \\
\bottomrule
\end{tabular*}
\end{table}

Table~\ref{tab:per_condition} gives the AUROC of every method in each test condition scored on its own, as the five-category mean; the per-condition AUROC and the worst condition (Section~\ref{sec:evaluation}) are in Table~\ref{tab:decomposition}. The single-condition AUROC need not decrease monotonically with severity, because normal and anomalous scores shift by different amounts at each level.

\section{Results per category}
\label{app:per_category}
\begin{table}[H]
\centering
\footnotesize\setlength{\tabcolsep}{4pt}
\caption{\textbf{Results per category.} Image-level logical AUROC (\%): clean (Cl.) and heterogeneous (Het.). Ours (no overlap): trained without test-overlapping augmentations.}
\label{tab:per_category}
\begin{tabular*}{\textwidth}{@{\extracolsep{\fill}}lcccccccccc@{}}
\toprule
 & \multicolumn{2}{c}{\shortstack{Breakfast\\Box}} & \multicolumn{2}{c}{\shortstack{Juice\\Bottle}} & \multicolumn{2}{c}{\shortstack{Pushpins}} & \multicolumn{2}{c}{\shortstack{Screw\\Bag}} & \multicolumn{2}{c}{\shortstack{Splicing\\Connectors}} \\
\cmidrule(lr){2-3}\cmidrule(lr){4-5}\cmidrule(lr){6-7}\cmidrule(lr){8-9}\cmidrule(lr){10-11}
Method & Cl. & Het. & Cl. & Het. & Cl. & Het. & Cl. & Het. & Cl. & Het. \\
\midrule
\textbf{Ours (no overlap)} & 92.8 & 93.3 & 92.7 & 92.3 & 99.5 & 99.0 & 91.3 & 90.1 & 89.5 & 87.8 \\
\textbf{Ours (full recipe)} & 97.1 & 96.4 & 97.0 & 96.4 & 100.0 & 99.8 & 90.1 & 87.7 & 90.3 & 90.2 \\
\midrule
CSAD & 95.8 & 60.0 & 95.5 & 71.2 & 99.7 & 66.2 & 99.9 & 87.3 & 97.4 & 56.1 \\
SALAD & 92.3 & 61.0 & 99.8 & 66.0 & 100.0 & 68.0 & 94.0 & 60.4 & 96.1 & 55.8 \\
LogSAD & 97.1 & 70.4 & 94.6 & 81.7 & 84.0 & 82.7 & 79.9 & 62.5 & 90.2 & 72.4 \\
PatchCore & 81.4 & 70.9 & 92.8 & 86.5 & 70.4 & 57.5 & 56.4 & 55.6 & 76.1 & 58.1 \\
\midrule
AnomalyCLIP & 68.5 & 61.9 & 61.7 & 60.2 & 54.6 & 50.6 & 47.8 & 48.7 & 53.7 & 50.1 \\
\bottomrule
\end{tabular*}
\end{table}

Table~\ref{tab:per_category} gives the clean and heterogeneous AUROC of every method per category. Our framework shows its largest drop in Splicing Connectors without test-overlapping augmentations and in Screw Bag with the full recipe; CSAD and SALAD show their largest drops in Splicing Connectors.

{\singlespacing
\bibliographystyle{elsarticle-harv}
\bibliography{main}
}

\end{document}